\documentclass[11pt,a4paper]{article}

\usepackage{arabtex}
\usepackage{times,latexsym}
\usepackage{url}
\usepackage[T1]{fontenc}
\usepackage{graphicx}
\usepackage{tabularx}
\usepackage{soul}
\usepackage{amsmath}
\usepackage{hyperref}
\usepackage{cleveref}

\usepackage{rotating}
\usepackage{tablefootnote}
\usepackage{cleveref}
 
\usepackage{adjustbox}
\usepackage{booktabs}
\usepackage{multirow}

\usepackage{microtype}
\usepackage[dvipsnames]{xcolor}
\usepackage{arabtex}
\usepackage{utf8}
\usepackage{xspace}
\usepackage{mathtools}
\usepackage{amssymb}
\usepackage{footnote}

\usepackage{graphicx}
\usepackage{color, colortbl}
\usepackage{xstring}
\usepackage{comment}
\usepackage{adjustbox}
\usepackage{float}
\restylefloat{table}
\usepackage{array,booktabs,makecell}
\usepackage{multirow}
\usepackage{graphicx}
\setcode{utf8}
\usepackage{appendix}
\usepackage{times,latexsym}
\usepackage{url}
\usepackage[T1]{fontenc}

\newcolumntype{H}{>{\setbox0=\hbox\bgroup}c<{\egroup}@{}}
\newcommand{\bluecheck}{{\color{blue}\checkmark}}

\usepackage{caption}
\usepackage{subcaption}
\usepackage{epstopdf}
\usepackage[utf8]{inputenc}

\usepackage{xstring}

\usepackage{color}
\usepackage{arydshln}
\usepackage{adjustbox}
\usepackage{float}
\restylefloat{table}
\usepackage{array,booktabs,makecell}
\usepackage{multirow}
\usepackage{graphicx}
\usepackage{color, colortbl}
\usepackage{xstring}

\usepackage{arydshln}

\usepackage{cancel}
\usepackage{ulem,xpatch}

\xpatchcmd{\sout}
  {\bgroup}
  {\bgroup}
  {}{}

\newcommand{\eat}[1]{}

\usepackage{caption}
\usepackage[font={footnotesize}]{caption}
\usepackage[]{acl2023}

\usepackage{xspace,mfirstuc,tabulary}

\crefname{section}{§}{§§}

\usepackage{arydshln}
\def \ourbench{ORCA}
\title{ \ourbench: A  Challenging Benchmark for Arabic Language Understanding}

\newcommand\blfootnote[1]{%
  \begingroup
  \renewcommand\thefootnote{}\footnote{#1}%
  \addtocounter{footnote}{-1}%
  \endgroup
}

\author{\normalsize AbdelRahim Elmadany$^{1,\star}$ ~ El Moatez Billah Nagoudi$^{1,\star}$ ~ Muhammad Abdul-Mageed$^{1,2,\star}$ \\
\normalsize $^{1}$ Deep Learning \& Natural Language Processing Group,
  The University of British Columbia\\\normalsize  $^{2}$Department of Natural Language Processing \& Department of Machine Learning, MBZUAI\\ %
  \texttt{\normalsize \{a.elmadany,moatez.nagoudi,muhammad.mageed\}@ubc.ca}}

\date{}
\begin{document}

\setcode{utf8}
\setarab 
\maketitle
\begin{abstract}
Due to the crucial role pretrained language models play in modern NLP, several benchmarks have been proposed to evaluate their performance. In spite of these efforts, no public benchmark of diverse nature currently exists for evaluating Arabic NLU. This makes it challenging to measure progress for both Arabic and multilingual language models. This challenge is compounded by the fact that any benchmark targeting Arabic needs to take into account the fact that Arabic is not a single language but rather a collection of languages and language varieties. In this work, we introduce a publicly available benchmark for Arabic language understanding evaluation dubbed ORCA. It is carefully constructed to cover diverse Arabic varieties and a wide range of challenging Arabic understanding tasks exploiting 60 different datasets (across seven NLU task clusters). To measure current progress in Arabic NLU, we use ORCA to offer a comprehensive comparison between 18 multilingual and Arabic language models. We also provide a public leaderboard with a unified single-number evaluation metric (\textit{ORCA score}) to facilitate future research.\footnote{\href{https://orca.dlnlp.ai/}{https://orca.dlnlp.ai/}.}



\end{abstract}

 \section{Introduction}\label{sec:intro}
 ~\blfootnote{ $^{\star}${All authors contributed equally.}}
 \begin{figure}[]
 \includegraphics[scale=0.18]{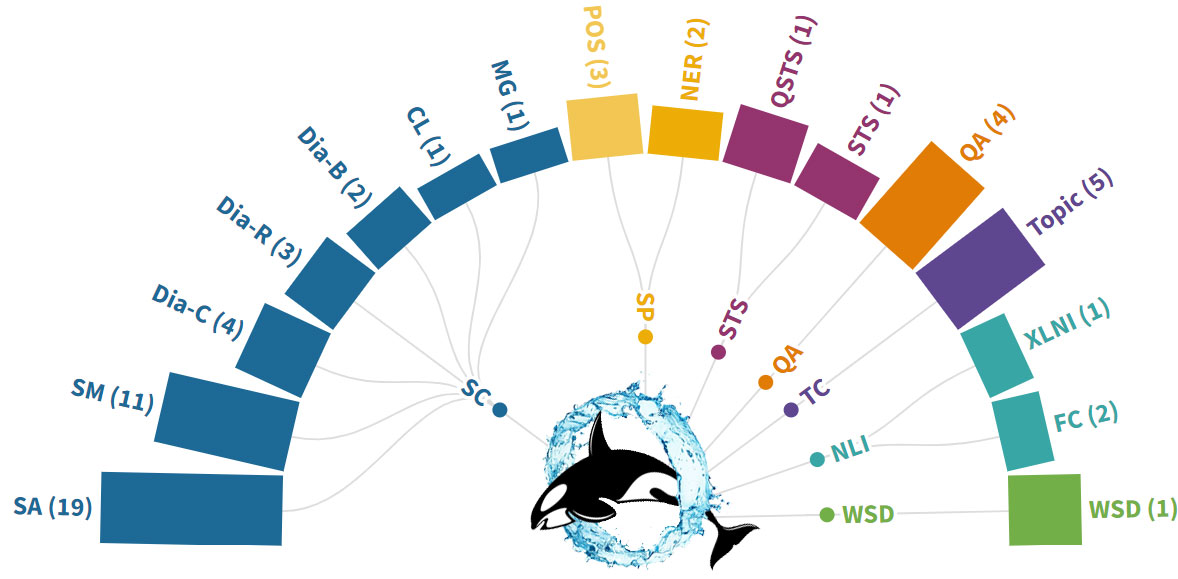}
  \caption{\ourbench~task clusters and datasets taxonomy.   The task clusters are \textbf{SC:} Sentence Classification.   \textbf{SP:} Structured Predictions.   \textbf{TC:} Topic Classification. \textbf{STS:}  Semantic Textual Similarity.  \textbf{NLI:} Natural Language Inference.  \textbf{QA:}  Question-Answering. \textbf{WSD:} Word sense disambiguation. The value in parentheses is the number of datasets in each  task cluster.   } \label{fig:sarlu_tax} 
 \end{figure}

\noindent  Pretrained language models (PLMs)~\cite{devlin2019bert,liu2019incorporating,lan2019albert, zhang-etal-2019-ernie, Sanh2019distilbert, radford2019gpt2, dai-etal-2019-transformer-xl, clark2020electra,lewis-etal-2020-bart, zaheer-2020-bigbard, brown-2020-gpt3, Beltagy2020LongformerTL, zhang-2020-pegasus, conneau-etal-2020-smlr, kitaev2020reformer, zhang-etal-2020-dialogpt, he2021deberta, pmlr-2021-lamda, raffel-t5-2022, Chung2022-flan-t5, Chowdhery2022PaLMSL} have become a core component of the natural language understanding (NLU) pipeline, making it all the more important to evaluate their performance under standardized conditions. For this reason, several English-based benchmarks such as GLUE~\cite{wang2018glue}, SuperGLUE~\cite{wang2019superglue}, SyntaxGym~\cite{gauthier-etal-2020-syntaxgym}, Evaluation Harness~\cite{eval-harness-2021}, GEM~\cite{gehrmann-etal-2021-gem}, NL-Augmenter~\cite{Dhole2021NLAugmenterAF}, Dynabench~\cite{kiela-etal-2021-dynabench}, MMLU~\cite{hendrycks2021mmlu}, NATURAL INSTRUCTIONS~\cite{mishra-etal-2022-cross}, BIG-bench~\cite{srivastava2022beyond}, and multilingual benchmarks such as XTREME~\cite{hu2020xtreme}, XGLUE~\cite{liang2020xglue}, and MASAKHANE~\cite{nekoto-etal-2020-masakhane} have been introduced. Benchmarks for a few other languages also followed, including FLUE~\cite{le2020flaubert} for French,  CLUE~\cite{xu2020clue} for Chinese,  IndoNLU~\cite{wilie2020indonlu} for Indonesian,  KorNLI and KorSTS~\cite{ham2020kornli} for Korean, and JGLUE\cite{kurihara2022jglue}. This leaves behind the majority of the world's languages, and relies on  multilingual benchmarks which often have limited coverage of dialects and naturally-occurring (rather than machine translated) text. This motivates us to introduce a benchmark for Arabic. One other reason that lends importance to our work is that Arabic is a rich collection of languages with both standard and dialectal varieties and more than $400$M native speaker population.



To the best of our knowledge, there have only been two attempts to provide Arabic NLU evaluation benchmarks. These are ARLUE~\cite{abdul2020arbert} and ALUE~\cite{seelawi2021-alue}. Although useful, both of these have major limitations: ALUE has \textit{modest coverage} (only eight datasets covering only three task clusters) and ARLUE involves datasets that are not publicly available. Our goal is to rectify these limitations by introducing \ourbench, which expands task coverage using \textit{fully} public datasets, while also offering an accessible benchmark with a public leaderboard and processing tools as well as wide geographical and linguistic coverage. \ourbench~exploits $60$ different datasets, making it by far the most extensive benchmark for Arabic NLU and among the most extensive for any language. We present detailed analyses of the data comprising \ourbench~and evaluate a wide range of available pretrained language models (PLMs) on it, thus offering strong baselines for future comparisons.

In summary, we offer the following contributions: \textbf{(1)} We introduce~\ourbench, an extensive and diverse benchmark for Arabic NLU. \ourbench~is  a collection of $60$ datasets arranged into \textit{seven  task clusters}, namely: sentence classification, text classification,  structured  prediction, semantic similarity, natural language inference (NLI), question-answering (QA), and word sense disambiguation (WSD). \textbf{(2)} We provide a comprehensive comparison of the performance of publicly available  Arabic PLMs on \ourbench~using a unified {\it ORCA score}. \textbf{(3)} To facilitate future work, we design a public leaderboard for scoring PLMs on~\ourbench. Our leaderboard is \textit{interactive} and offers \textit{rich meta-data} about the various datasets involved as well as the language models we evaluate. \textbf{(4)} We distribute~\ourbench~with a new \textit{modular toolkit} for pretraining and  transfer learning for NLU. The toolkit is built around standard tools including PyTorch \cite{paszke2019pytorch} and HuggingFace datasets hub~\cite{lhoest2021datasets}.

The rest of the paper is organized as follows: In Section~\ref{sec:lit}, we provide an overview of related work. Section~\ref{sec:SARLUE} introduces \ourbench, our Arabic NLU benchmark. In Section~\ref{sec:LM}, we describe multilingual and Arabic pretrained language models we evaluate on~\ourbench, providing results of our evaluation in Section~\ref{sec:Eval}. Section~\ref{sec:diss} is an analysis of model computational cost as measured on~\ourbench. We conclude in Section~\ref{sec:conc}.

\section{Related Work}\label{sec:lit}

Most recent benchmarks propose a representative set of standard NLU tasks for evaluation. These can be categorized into English-centric, multilingual,  Arabic-specific, and X-specific (X being a language other than English or Arabic such as Chinese, Korean, or French). We briefly describe each of these categories next. We also provide a comparison of benchmarks in the literature in terms of task clusters covered and number of datasets in Table~\ref{tab:bench_comp}.

\subsection{English-Centric Benchmarks} 
\textbf{GLUE.} The General Language Understanding Evaluation (GLUE) benchmark~\cite{wang2018glue} is one of the early English benchmarks. It is a collection of nine publicly available datasets from different genres. GLUE is arranged into three task clusters: sentence classification, similarity and paraphrase, and NLI. 

\noindent\textbf{SuperGLUE.}~\newcite{wang2019superglue} propose SuperGLUE, a benchmark styled after GLUE with a new set of more challenging tasks. SuperGLUE is built around eight tasks and  arranged into four task clusters: QA, NLI, WSD, and coreference resolution. The benchmark is accompanied by a leaderboard with a single-number performance metric (i.e., the \textit{SuperGLUE score}). 


\subsection{Multilingual Benchmarks} 

\begin{table*}[] 
\centering
 \renewcommand{\arraystretch}{1.5}
\resizebox{1\textwidth}{!}{%


\begin{tabular}{lc|cc|Hccccc|ccc|ccc}\toprule

\multirow{3}{*}{\textbf{Task Cluster}} &&\multicolumn{2}{|c}{\textbf{English-Centric}}&\multicolumn{6}{|c}{\textbf{X-Specific} }&\multicolumn{3}{|c}{\textbf{Multilingual} }&\multicolumn{3}{|c}{\textbf{Arabic} } \\
\cline{2-16}
&\textbf{Bench.} & \textbf{GLUE} & \textbf{SGLUE} &  \textbf{XGLUE} &  \textbf{FLUE} & \textbf{IndoNLU} & \textbf{CLUE} & \textbf{JGLUE} &\textbf{KorNLU} &\textbf{bAbI} &  \textbf{XGLUE} & \textbf{XTREM} & \textbf{ALUE} & \textbf{ARLUE} & \colorbox{green!30}{\textbf{\ourbench}}\\

\cline{2-16}
   &  \textbf{Lang.}&  En        & En  & 19    & Fr        & Id           & Zh        & Jp & Ko & En, Hi & 19     & 40         & Ar        & Ar         & Ar          \\  \toprule

Sentence Classification &  & \bluecheck      &    &      & \bluecheck      & \bluecheck        & \bluecheck   & \bluecheck  &   &         &      & \bluecheck      & \bluecheck     & \bluecheck      & \bluecheck       \\

Structured Prediction    &   &      & \bluecheck  & \bluecheck      & \bluecheck     & \bluecheck        &  &       &  &     & \bluecheck        & \bluecheck      &          & \bluecheck      & \bluecheck       \\

STS and Paraphrase  &  & \bluecheck     &    & \bluecheck       & \bluecheck     & \bluecheck        & \bluecheck    &       & \bluecheck                 &      & \bluecheck      &            & \bluecheck     &            & \bluecheck       \\

Text Classification    &  & \bluecheck     &     &         &           &              & \bluecheck    &\bluecheck       &    &           &   &                 &           & \bluecheck      & \bluecheck      \\

Natural Language Inference   &  &                 & \bluecheck  & \bluecheck     & \bluecheck    & \bluecheck           & \bluecheck  &      & \bluecheck                 &         & \bluecheck   &            & \bluecheck     &            & \bluecheck       \\

Word  Sense  Disambiguation  &  &           & \bluecheck     &   & \bluecheck     &              &        &       &   &       &       &            &           &            &      \bluecheck        \\

Coreference Resolution    &  & \bluecheck     & \bluecheck    &    &           &              & \bluecheck    &       &   & \bluecheck     &   &            &           &            &             \\

Question-Answering      &  & \bluecheck     & \bluecheck     &   &           & \bluecheck        & \bluecheck    &  \bluecheck       &   & \bluecheck  &     & \bluecheck      &           & \bluecheck      & \bluecheck    \\ \toprule   

\textbf{\# Datasets}     &  &    11 & 10    & 11   &  7    &     12      & 9      & 6      & 4   &   20  & 11      & 9      &  9         &  42     & \textbf{60}    \\    \toprule

\textbf{\# Task Clusters Covered}     &  &    5 & 5    & 3   & 5     &     5      & 6       &3    &  2    &   2 & 3    & 4     &  3         & 4      & \textbf{7}    \\    \toprule


%

\end{tabular}}
\caption{Comparison of  NLU benchmarks proposed in the literature across the different covered task clusters. \textbf{STS}: Semantic Textual  Similarity. \textbf{GLUE}: \cite{wang2018glue}.  \textbf{SGLUE}: SuperGLUE ~\cite{wang2019superglue}.  \textbf{XGLUE}: \cite{liang2020xglue}. \textbf{FULE}: \cite{le2020flaubert}. \textbf{FULE}: \cite{le2020flaubert}.  \textbf{IndoNLU}: \cite{wilie2020indonlu}. \textbf{CLUE}:~\cite{xu2020clue}.  \textbf{KorNLI}: KorNLI  and korSTS~\cite{ham2020kornli}. \textbf{bAbI}: \cite{weston2015towards}. \textbf{XTREM}: \cite{hu2020xtreme}. \textbf{ALUE}:~\cite{seelawi2021-alue}. \textbf{ARLUE}: \cite{abdul2020arbert}. \colorbox{green!20}{\textbf{\ourbench:}}  Our proposed Arabic NLU benchmark.}
\label{tab:bench_comp}

\end{table*}


\noindent\textbf{bAbI.} Early attempts to create multilingual benchmarks are limited in their language coverage. An example is bAbI~\cite{weston2015towards}, which covers only English and Hindi. It consists of a set of $20$ tasks for testing text reasoning and understanding  using different question-answering and coreference resolution strategies. 

\noindent\textbf{XGLUE.}
XGLUE is a cross-lingual benchmark proposed by \newcite{liang2020xglue} to evaluate the performance of PLMs. It  provides $11$~tasks in both NLU and NLG scenarios that cover $19$ languages. The XGLUE tasks are arranged into four \textit{understanding tasks} (structured predictions, text classifications,  QA,  NLI, semantic search), and two \textit{generation tasks} (question and title generation).


\noindent\textbf{XTREME.} The Cross-lingual TRansfer Evaluation of Multilingual Encoders  (XTREME)~\cite{hu2020xtreme} is a benchmark for evaluating the cross-lingual generalization capabilities of multilingual models. It covers $40$ languages and includes nine datasets across four task clusters: classification (i.e., NLI and paraphrase), structured prediction (i.e., POS and NER), question answering, and sentence retrieval. 
~\newcite{ruder-etal-2021-xtreme-r} extend XTREME to XTREME-R (for XTREME Revisited). This new benchmark has an improved set of ten NLU tasks (including language-agnostic retrieval tasks). XTREME-R covers $50$ languages. Authors also provide a multilingual diagnostic suite and evaluation capabilities through an interactive public leaderboard.

\noindent\textbf{Big-bench.} The Beyond the Imitation Game Benchmark or shortly BIG-bench~\cite{srivastava2022beyond}  is a collaborative\footnote{Contributed by
$444$ authors across $132$ institutions.} NLP benchmark  aimed to explore and evaluate the capabilities of large language models. It currently consists of $204$ advanced NLP tasks, from diverse topics such as common-sense reasoning, linguistics, childhood development, math, biology, physics, social bias, and software development.\footnote{We exclude the Big-Bench benchmark from
Table~\ref{tab:bench_comp} because it has a very large number of tasks that we cannot fit into the table. It also involves task clusters that are unrelated to the~\ourbench~benchmark.}

\subsection{Arabic-Specific Benchmarks} 

\noindent\textbf{ALUE.} To the best of our knowledge, two benchmarks for Arabic currently exist, ALUE~\cite{seelawi2021-alue} and ARLUE~\cite{abdul2020arbert}. ALUE is focused on NLU and comes with eight datasets arranged into three task clusters: sentence classification, NLI, and similarity and paraphrase. The sentence classification cluster involves five datasets for offensive and hate speech detection, irony prediction, sentiment analysis, and dialect identification. The NLI cluster involves two datasets, both for NLI aiming at predicting whether a premise sentence contradicts, entails, or is neutral toward a hypothesis sentence. ALUE has one dataset for semantic similarity comprising a collection of questions pair labelled  with "1" (semantically similar) or "0" otherwise. The task is to predict these similarity scores. While datasets in ALUE are publicly available and the benchmark is accompanied by a leaderboard, its size and diversity (geographical and linguistic) are modest.\\
\noindent\textbf{ARLUE.}~\cite{abdul2020arbert} also targets Arabic NLU tasks and is composed of $42$ datasets arranged into four task clusters: sentence classification, text classification, structured prediction, and QA. Many of the datasets in ARLUE, however, are not publicly available which presents a barrier to widespread adoption. Nor is ARLUE accompanied by a leaderboard.\textit{~\ourbench~ameliorates these challenges.} \\

\subsection{X-Specific Benchmarks} 

Other X-specific benchmarks include \textbf{CLUE.}~\cite{xu2020clue}, \textbf{FLUE.}~\cite{le2020flaubert}, \textbf{IndoNLU.}~\cite{wilie2020indonlu}, \textbf{JGLUE.}~\cite{kurihara2022jglue}, and \textbf{KorNLI and KorSTS.}~\cite{ham2020kornli}. We review these benchmarks in Appendix~\ref{appdx_sec:xs_benchmark}.


\section{\ourbench~Benchmark}\label{sec:SARLUE}

We present~\ourbench, a benchmark for Arabic NLU that is challenging and diverse.~\ourbench~ involves $60$ datasets arranged into $29$ tasks and seven task clusters. In the following, we will first introduce our design principles for developing~\ourbench~then introduce the different task clusters covered.


\begin{table}[ht]
\centering
 \renewcommand{\arraystretch}{1.18}
\resizebox{\columnwidth}{!}{%
\begin{tabular}{lllHHHHcrrr}

\toprule
\multirow{1}{*}{\textbf{Cluster}} &  \multirow{1}{*}{\textbf{Task}}  & \multirow{1}{*}{\textbf{Level}}  & \textbf{\#Data}&\textbf{Train}   & \textbf{Dev} & \textbf{Test}  & \textbf{\#Data} &\textbf{Train}   & \textbf{Dev} & \textbf{Test}                    \\ \toprule 
 \multirow{7}{*}{\textbf{SC}}  &\colorbox{red!0}{SA} & Sent & $17$&  $190.9$K & $6.5$K  & $50$K & $19$  & $50$K  & $5$K& $5$K   \\  
  &\colorbox{red!0}{SM~~~} & Sent &  $8$ &  $1.51$M  &  $162.5$K  & $166.1$K & $11$  & $50$K  & $5$K & $5$K   \\  
 &\colorbox{orange!0}{Dia-b} &Sent &  $2$ &  $94.9$K & $10.8$K  & $12.9$K& $2$   & $50$K  & $5$K& $5$K   \\  
 &\colorbox{red!0}{Dia-r} &Sent &  $2$ &  $38.5$K & $4.5$K  & $5.3$K& $3$   & $38.5$K  & $4.5$K& $5$K   \\  
  &\colorbox{red!0}{Dia-c}  &Sent & $3$  &  $711.9$K & $31.5$K  & $52.1$K& $4$   & $50$K  & $5$K& $5$K   \\ 
  & \colorbox{green!0}{CL} & Sent &  - &  - & -  & - & $1$  & $3.2$K  & $0.9$K& $0.4$K   \\ 
  & \colorbox{green!0}{MG} &Sent &   - &  - & -  & - & $1$ & $50$K  & $5$K& $5$K   \\   \hline  
 \multirow{2}{*}{\textbf{SP}} & \colorbox{orange!0}{NER} & Word &  $5$&  $277.7$K& $44.1$K & $66.5$K& $2$ & $5.2$K  & $1.1$K& $1.2$K   \\   
 & \colorbox{orange!0}{POS} & Word &  $1$&  $277.7$K& $44.1$K & $66.5$K& $2$ & $5.2$K  & $1.1$K& $1.2$K   \\   \hline  
\textbf{TC} &\colorbox{orange!0}{Topic}&Doc & $5$  &  $47.5$K & $5.9$K  & $5.9$K& $5$ & $47.5$K  & $5$K& $5$K   \\   \hline 

\textbf{QA} & \colorbox{orange!0}{QA} &Parag &  $4$&  $101.6$K& $0.5$K & $7.4$K & $4$  & $101.6$K& $517$ & $7.4$K \\  \hline
\multirow{2}{*}{\colorbox{blue!0}{\textbf{STS}}} &\colorbox{green!0}{STS-reg} & Sent &   - &  - & -  & - & $1$ & $0.8$K  & $0.2$K& $0.2$K   \\  
 &\colorbox{green!0}{STS-cls} & Sent &  - &  - & -  & - & $1$ & $9.6$K  & $1.2$K& $1.2$K   \\ \hline 
\multirow{2}{*}{\colorbox{blue!0}{\textbf{NLI}}}  &\colorbox{green!0}{XNLI}&Sent &  - &  - & -  & - & $1$  & $4.5$K  & $0.5$K& $2.5$K   \\  

 & \colorbox{green!0}{FC} & Doc  &  - &  - & -  & - & $2$  & $5$K  & $1$K& $1$K   \\ \hline
\textbf{WSD} &\colorbox{blue!0}{WSD}& Word &  $5$  &  $47.5$K & $5.9$K  & $5.9$K& $1$ & $21$K  & $5$K& $5$K   \\   \hline 
\textbf{Total}  &  &  &  42 &  $2.9$M & $256$K  & $366$K  & $\bf60$ & $487.1$K  & $46.0$K& $55.1$K   \\ 
\midrule
\end{tabular}%
 }
\caption{\small{The different task clusters, tasks, and data splits in~\ourbench. \textbf{SC:} Sentence Classification.   \textbf{SP:} Structured Prediction.   \textbf{TC:} Topic Classification. \textbf{STS:} Textual Semantic Similarity.  \textbf{NLI:} Natural Language Inference.  \textbf{QA:}  Question-Answering.  \textbf{SM}: Social Meaning. For abbreviations of task names, refer to Section~\ref{subsec:orca-tasks}. }}
\label{tab:orca-data}
\end{table}
\subsection{Design Principles}

Our goal is to offer a \textit{challenging} and \textit{diverse} NLU benchmark that allows evaluation of language models and measurement of progress on Arabic NLU. To this end, we develop~\ourbench~with a number of design principles in mind. We explain these here.

\noindent\textbf{Large number of public tasks.} We strive to include as many tasks as possible so long as data for these tasks are public. This makes it possible for researchers to train and evaluate on these tasks without having to pay for private data.~\ourbench~ involves $60$ different datasets that are \textit{all} publicly available.  


 
\begin{figure*}
     \centering
     \begin{subfigure}[b]{\columnwidth}
         \centering
         \includegraphics[width=\textwidth]{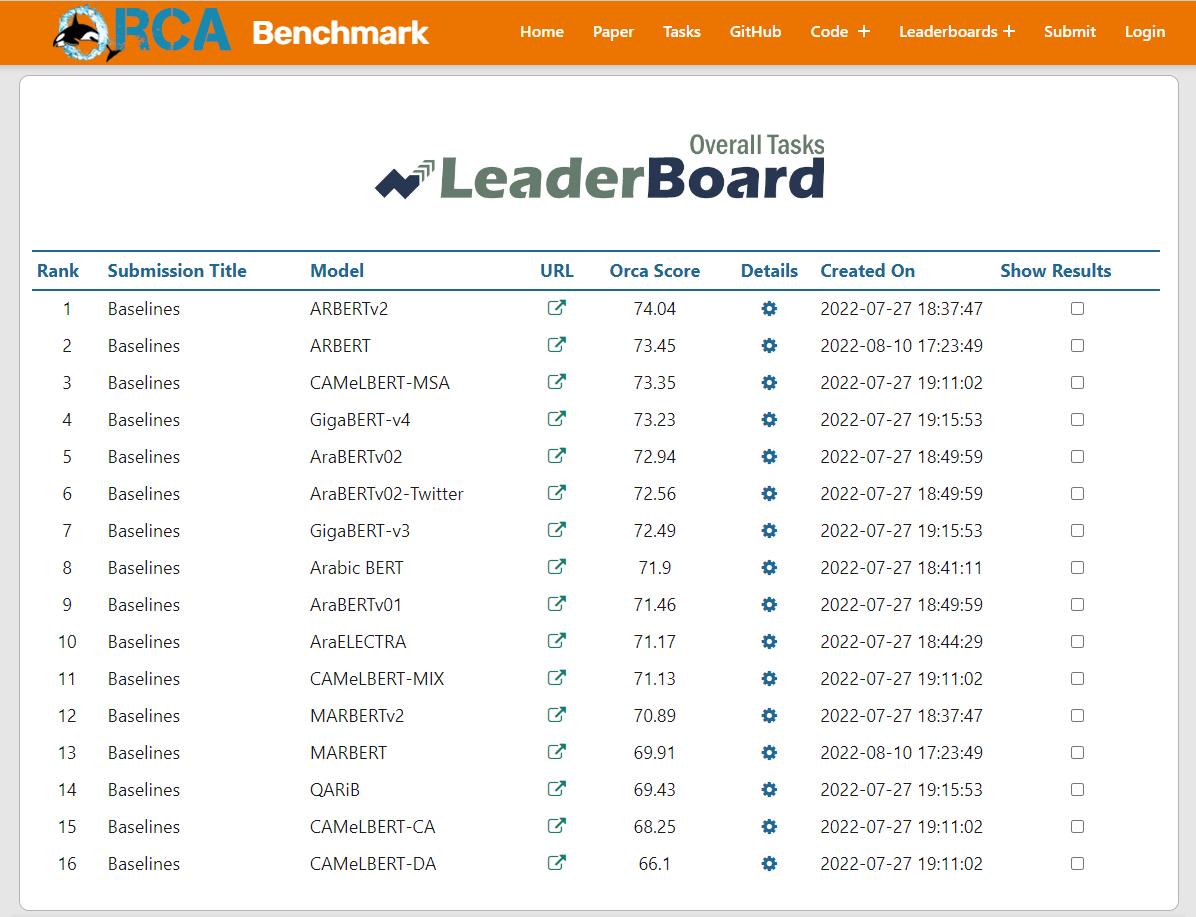}
         \caption{Models ranked by our ORCA score.}
         \label{fig:orca_main_leaderboard_a}
     \end{subfigure}
     \hfill
     \begin{subfigure}[b]{\columnwidth}
         \centering
         \includegraphics[width=\textwidth]{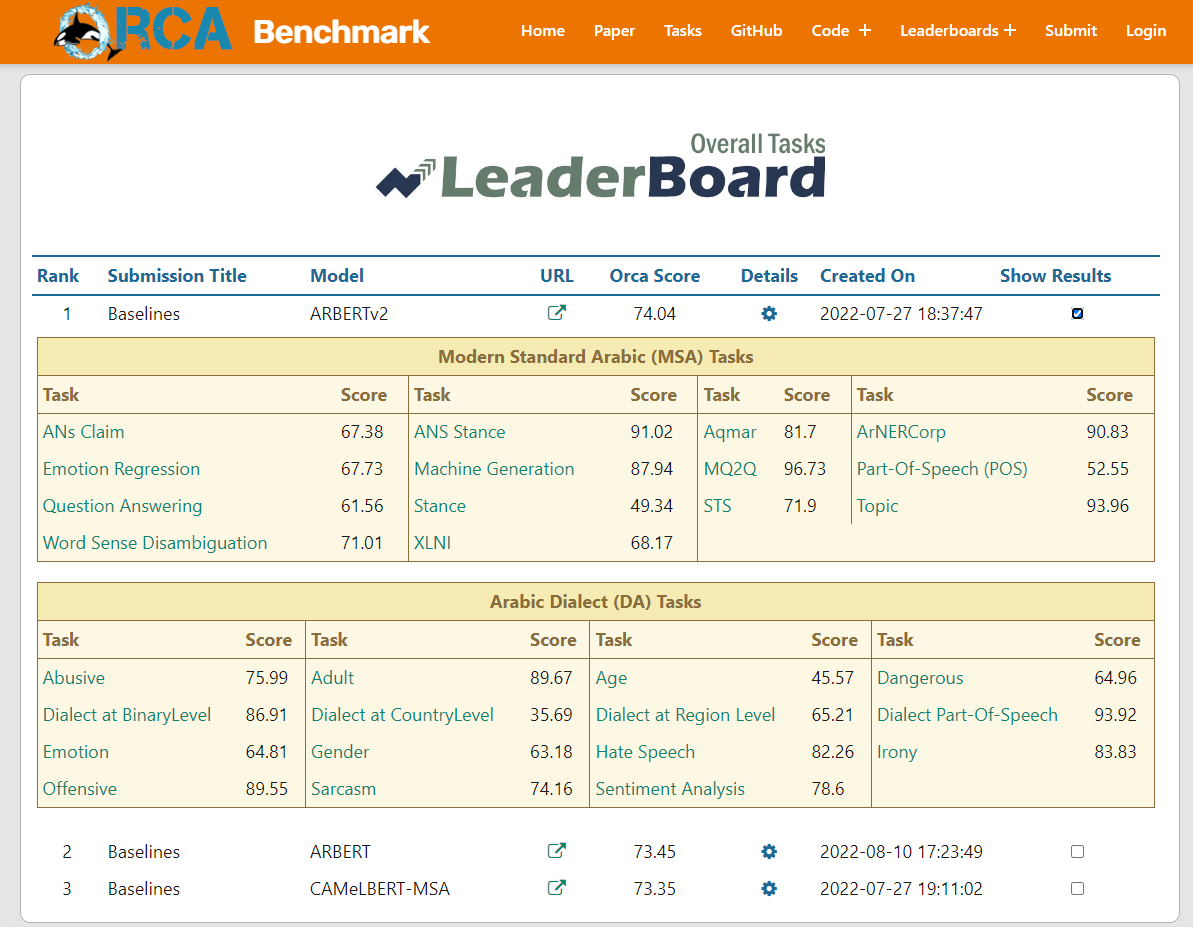}
         \caption{Detailed scores for a given model across all tasks.}
         \label{fig:orca_main_leaderboard_b}
     \end{subfigure}
     \caption{ORCA main leaderboard.}
     \label{fig:orca_main_leaderboard}
\end{figure*}
\noindent\textbf{Challenging benchmark.}
We design~\ourbench~ to require knowledge at various linguistic levels, making it challenging. This includes knowledge at the level of tokens in context as well as at the levels of complete sentences, inter-sentence relations, whole paragraphs, and entire documents.




\noindent\textbf{Coherent task clusters and tasks.} Rather than listing each group of datasets representing a given task together, we group the various tasks into \textit{task clusters}. This makes it easy for us to present the various downstream tasks. It also makes it possible to derive meaningful insights during evaluation. For example, one can compare performance at a lower-level task cluster such as structured prediction to that of performance at a higher-level cluster such as natural language inference. Within the clusters themselves, we also maintain coherent subgroupings. For example, since sentiment analysis has been one of the most popular tasks in Arabic NLP, we assign it its own sub-cluster within the sentence classification cluster. Similarly, we keep tasks such as hate speech and emotion detection that exploit social media data into a single {\it social meaning} cluster.


\noindent\textbf{Wide linguistic variability and geographical coverage.} We strive to include tasks in various Arabic varieties. This involves  Modern Standard Arabic (MSA) and dialectal Arabic (DA). We include datasets collected from wide regions within the Arab world. This not only pertains our DA datasets, many of which come from various Arab countries, but also our MSA datasets as these are extracted from several news outlets from across the Arab world. This also ensures variability in topics within these datasets. To illustrate linguistic diversity within~\ourbench, we run an in-house binary MSA-dialect classifier on all~\ourbench~data splits (i.e., Train, Dev, and Test).\footnote{As our classifier is trained using~\ourbench\textsubscript{DA}, we exclude the~\ourbench~dialect component from this analysis.}  For a deeper understanding of~\ourbench~data, we also calculate several measures including the average, median,  mode, and  type-token ratio (TTR) of the sentences in each task.  Table~\ref{tab:SuperARLUE_msa_dia} shows the MSA vs. DA data distribution and the statistical description of~\ourbench~datasets.

\begin{table*}[t]
\centering
 \renewcommand{\arraystretch}{1}
\resizebox{1\textwidth}{!}{%
\begin{tabular}{llHccccccc}
\toprule

 \multicolumn{9}{c}{\colorbox{purple!15}{\textbf{Task cluster with more likely dialectal data}}} \\
\toprule 
\textbf{Cluster} & \textbf{Task}  & \textbf{\#Datasets}&\textbf{Avg-char}   & \textbf{Avg-word} & \textbf{Median}  & \textbf{Mode} &\textbf{TTR}   & \textbf{MSA} & \textbf{DIA}                    \\ \toprule

&abusive&$xx$&$43.71$&$12.45$&$11$&$8$&$27.69$&$29.55$&$70.45$\\
&adult&$xx$&$86.15$&$15.44$&$14$&$3$&$18.65$&$65.8$&$34.2$\\
&age$^\star$&$xx$&$60.73$&$11.82$&$15$&$19$&$42.44$&$41.52$&$58.48$\\
&{claim}&$xx$&$48.23$&$8.16$&$8$&$7$&$37.96$&$99.78$&$0.22$\\	
\multirow{9}{*}{\textbf{SC}}&dangerous&$xx$&$38.27$&$8.17$&$8$&$7$&$35.69$&$10.16$&$89.84$\\
&{dialect-B}&$xx$&$89.92$&$17.17$&$10$&$4$&$37.84$&$60.27$&$39.73$\\
&{dialect-C}&$xx$&$82.66$&$15.39$&$17$&$22$&$37.62$&$27.44$&$72.56$\\
&{dialect-R}&$xx$&$80.40$&$15.66$&$8$&$8$&$36.32$&$12.57$&$87.43$\\	
&emotion-cls&$xx$&$73.58$&$14.60$&$16$&$10$&$14.25$&$25.72$&$74.28$\\
&emotion-reg&$xx$&$63.57$&$12.60$&$14$&$9$&$14.25$&$60.3$&$39.70$\\
&gender$^\star$&$xx$&$60.73$&$11.82$&$9$&$4$&$42.44$&$40.9$&$59.10$\\
&hate$^\dagger$&$xx$&$99.20$&$19.76$&$16$&$9$&$24.97$&$25.79$&$74.21$\\
&offensive$^\dagger$&$xx$&$99.20$&$19.76$&$16$&$9$&$24.97$&$25.79$&$74.21$\\
&irony&$xx$&$106.67$&$19.70$&$18$&$17$&$31.15$&$45.32$&$54.68$\\	
&{machine G}&$xx$&$218.11$&$39.92$&$33$&$31$&$14.44$&$99.44$&$0.56$\\	
&sarcasm&$xx$&$88.80$&$15.69$&$16$&$18$&$28.29$&$71.49$&$28.51$\\	
&{sentiment}&$xx$&$127.27$&$22.9$&$16$&$10$&$79.31$&$64.06$&$35.94$\\
\hdashline
 \textbf{\colorbox{cyan!10}{\textbf{Avg}}} &   & xx&$86.31$ & $16.53$ & $14.41$ & $11.47$ & $32.25$ & \colorbox{green!15}{$47.41$} & \colorbox{purple!15}{$52.59$}\\
	



\toprule 
 \multicolumn{9}{c}{\textbf{\colorbox{green!15}{\textbf{Task clusters with more likely MSA data}}}} \\
\toprule 

\textbf{TC}& topic&$xx$&$2.7k$&$474.78$&$286$&$152$&$5.2$&$99.71$&$0.29$\\  
\textbf{QA}&arlue-qa&$xx$&$$101.6$$&$$517$$&$$7.4$$&$$4$$&$$50$$&$100$&$0.0$\\	  \hdashline

  \textbf{\colorbox{cyan!10}{\textbf{ One input avg }}} &  &xx&$1.4$k & $495.89$ & $146.7$ & $78$ & $27.6$ & \colorbox{green!15}{$99.86$} & \colorbox{purple!15}{$0.15$}\\				
\hline  
 \multirow{3}{*}{{\textbf{NLI}}}  
& ans-st&$xx$&$50.53/45.35$&$8.48/7.70$&$44780$&$44749$&$36.46/38.70$&$99.85$&$0.15$\\	
& baly-st&$xx$&$7.2k/147.65$&$1.3k/25.40$&$25/807$&$8.12/5.24$&$21/251$&$100$&$0.0$\\
& xlni&$xx$&$90.12/44.15$&$16.23/7.97$&$15/7$&$9/7$&$13.74/28.76$&$99.16$&$0.84$\\	

\multirow{2}{*}{{\textbf{STSP}}} 
& {sts-reg}&$$4$$&$78.72/96.44$&$14.19/17.26$&$14/13$&$12/8$&$60.07/58.13$&$98.86$&$1.14$\\	
&{sts-cls}&$xx$&$80.38/77.25$&$14.25/13.13$&$11/10$&$7/7$&$10.03/10.33$&$99.31$&$0.69$\\ \hdashline

\textbf{\colorbox{cyan!10}{\textbf{Two inputs avg}}} & &$xx$& $1.5$k/$82.17$ & $270.63/14.29$ & $171/12.4$ & $8.62/6.65$ & $28.26/77.38$ & \colorbox{green!15}{$99.44$} & \colorbox{purple!15}{$0.56$}\\\hline
\toprule

\end{tabular}%
}
\caption{\small{Descriptive statistics of~\ourbench~across the different data splits. $^\star$ and $^\dagger$: Same data with multiple labels. \textbf{SC:}~Sentence Classification.      \textbf{TC:} Topic Classification. \textbf{STS:} Textual Semantic Similarity.  \textbf{NLI:}~Natural Language Inference.  \textbf{QA:}  Question-Answering. For the NLI and STPS tasks we compute the statistics in both inputs (e.g., sentence 1 and sentence 2 in ASTS task). We don't include the word-level datasets in this table (i.e., SP tasks.)  }  }
\label{tab:SuperARLUE_msa_dia}
\end{table*}

In addition, we acquire a country-level dialect distribution analysis over the data using AraT5 model~\cite{nagoudi2021arat5} fine-tuned on the \ourbench~dialect country-level dataset (DIA-C). We run this country-level classifier only on the dialectal portion of \ourbench~(i.e., datasets of tweets predicted as \textit{dialectal} with our in-house MSA-dialect classifier). Figure~\ref{fig:map_countery} shows that  \ourbench~datasets are \textit{truly diverse} from a geographical perspective.\footnote{Again, the country-level classifier is also trained using~\ourbench\textsubscript{DIA}, so we exclude the dialect tasks from this analysis.}

\noindent\textbf{Accessible evaluation.} To facilitate evaluation in a reasonable time frame in a GPU-friendly setting, we cap data sizes across our Train, Dev, and Test splits to \textbf{\textit{50k, 5k, 5k }} samples respectively. This allows us to avoid larger data sizes in tasks such as \textit{Age}  and \textit{Gender} that have \textit{1.3m, 160k, 160k} samples for the Train, Dev, and Test splits each and both the \textit{sentiment} and \textit{dialect country-level} tasks  that have \textit{190k, 6.5k, 44.2k} and \textit{711.9k, 31.5k,  52.1k} for the Train, Dev, and Test data (respectively). Table~\ref{tab:orca-data}, shows a summary of the data splits across tasks and task clusters in~\ourbench.

\noindent\textbf{Simple evaluation metric.} We adopt a simple evaluation approach in the form of an {\it \ourbench~score}. The \ourbench~score is simply a macro-average of the different scores across all tasks and task clusters, where  each task is weighted equally.

\noindent\textbf{Modularity.} We design~\ourbench~ to allow users to score models on the whole benchmark but also on individual task clusters. In both cases, the leaderboard returns results averaged across the datasets within either the whole benchmark or the individual tasks (sub-leaderboards). This allows us to invite submissions of dedicated models that take as its target subsets of~\ourbench~datasets. Figure~\ref{fig:orca_main_leaderboard} shows ORCA's main screen with models sorted by ORCA score. We provide more screenshots illustrating ~\ourbench's modularity in Appendix~\ref{apdx:leaderboard}.


\noindent\textbf{Public leaderboard.}
We allow scoring models against~\ourbench~through an intuitive, easy-to-use leaderboard. To facilitate adoption, we also provide a Google Colab notebook with instructions for finetunining any model on~\ourbench~tasks.

\noindent\textbf{Faithful evaluation.}
For each submission, we require users to provide meta-data such as the number of parameters, amount of pretraining data, and number of finetuning epochs. This facilitates comparisons across the different models. We make this meta-data available via our interactive leaderboard. 

\noindent\textbf{Proper credit for individual dataset authors.} One issue with evaluation benchmarks is that once a benchmark is created there is a concern of not giving credit to original datasets. To overcome this limitation, we distribute a simple text file with bibliographic entries for all papers describing the $60$ datasets in~\ourbench~ and strongly encourage all future use to cite them.

\subsection{Tasks and Task Clusters}\label{subsec:orca-tasks}


As explained, we arrange \ourbench~into $7$ task clusters. These are (1) sentence classification, (2) structured prediction (3) semantic textual  similarity and paraphrase, (4) text classification, (5) natural language inference, (6) word sense disambiguation, and (7) question answering.

\noindent\textbf{Sentence Classification.}
This cluster involves the following sentence-level classification tasks:~\textit{(1) Sentiment Analysis:} $19$ publicly available sentiment datasets have been used to construct this task. We merge $17$ datasets benchmarked by~\newcite{abdul2020arbert} with two new datasets:  Arabizi sentiment analysis dataset \cite{Chayma2020} and AraCust~\cite{almuqren2021aracust}, a Saudi Telecom tweets corpus for sentiment analysis. \textit{(2)~Social Meaning:} Refers to eight social meaning datasets covering prediction of hate speech and offensive language \cite{mubarak-etal-2020-overview}, dangerous speech~\cite{alshehri2020understanding}, sarcasm~\cite{farha2020arabic}, adult content~\newcite{mubarak2021adult}, irony~\cite{idat2019}, emotion, age and gender~\cite{mohammad2018semeval,mageed-2020-aranet}. \textit{(3)  Dialect Identification:} Six datasets are used for dialect classification. These are ArSarcasm\textsubscript{Dia}~\cite{farha2020arabic}, the Arabic Online Commentary (AOC) dataset ~\cite{zaidan2014arabic}, NADI-2020~\cite{mageed-etal-2020-nadi}, MADAR~\cite{bouamor2019madar},   QADI~\cite{abdelali2020arabic}, and Habibi~\cite{el2020habibi}. The  dialect identification task involves three  dialect classification levels. These are the binary-level (i.e., MSA vs. DA), region-level  (four regions), and  country-level  ($21$ countries). \textit{(4) Claim Prediction}: we  use ANS-claim~\cite{khouja2020stance}, which is a  factuality prediction of claims corpus created using the credibility of the editors as a proxy for veracity (true/false). \textit{(5) Machine Generation:} for machine generated text detection (i.e, machine vs. human), we use the machine manipulated version of AraNews dataset~\cite{nagoudi2020machine}. To create this dataset, a list of words are selected (based on their POS tags) and substituted by a chosen word from the $k$-most similar words in an Arabic word embedding model.

\noindent\textbf{Structured Prediction}. This task cluster includes two tasks: \textit{(1) Arabic NER}: we  consider  two publicly available Arabic NER datasets, ANERcorp~\cite{benajiba2007anersys} and AQMAR~\cite{schneider2012coarse}. \textit{(2) Arabic POS Tagging}: we use two POS Tagging datasets,  the multi-dialect Arabic POS dataset~\newcite{darwish2018multi} and the Arabic POS tagging part of XGLUE~\cite{liang2020xglue}.

\noindent\textbf{Text Classification.} In this task cluster, we explore \textit{topic classification} employing three document-level classification datasets:  Khaleej~\cite{abbas2011evaluation}, Arabic News Text (ANT)~\cite{chouigui2017ant}, and OSAC~\cite{saad2010osac}. %
 
\noindent\textbf{Semantic Textual Similarity.} This cluster  aims to measure the semantic relationship between a pair of sentences. For this, we use the  \textit{(1) STS regression: } data from Ar-SemEval-2017~\cite{cer2017semeval} (which is a set of Arabic sentence pairs each labeled with a numerical score from the interval $[0..1]$ indicating the degree of semantic similarity).  We also use \textit{(2) STS classification} where a pair of questions is assumed to be semantically similar if they have the same exact meaning \textit{and} answer. We use  the semantic question similarity in Arabic dataset (Q2Q) proposed by~\newcite{seelawi2019nsurl} where  each pair is tagged with ``1" (question has the same meaning and answer) or ``0" (not similar).

 \noindent\textbf{Natural Language Inference.} This cluster covers the following two tasks: \textit{(1) Arabic NLI:} we use the Arabic part of the cross-lingual natural language inference (XNLI) corpus~\cite{conneau2018xnli}. The goal is determining whether a text (hypothesis) is false (contradiction), undetermined (neutral), or true (entailment), given a another text (premise).  \textit{(2) Fact-checking}: in order to build a  fact-checking benchmark component, we use Unified-FC~\cite{baly2018integrating} and ANS~\cite{khouja2020stance}. Both of these datasets target stance and factuality prediction of claims from news and social media. The two datasets are manually created by annotating the stance between a claim-document pair with labels from the set \textit{\{agree, disagree, discuss, unrelated\}}.

\noindent\textbf{Word Sense Disambiguation.} We use the Arabic WSD  benchmark~\cite{el2021arabic}, a context-gloss pair dataset extracted from an MSA dictionary. It consists of $15$k senses for $5$k unique words with an average of three senses for each word.

\noindent\textbf{Question Answering.} We concatenate four Arabic and multilingual  QA datasets. These are ARCD \cite{mozannar2019neural}, MLQA~\cite{lewis2019mlqa}, TyDi QA~\cite{artetxe2020cross}, and XQuAD~\cite{artetxe2020cross}.

\begin{table*}[ht] 

\centering
 \renewcommand{\arraystretch}{1.25}
\resizebox{1\textwidth}{!}{%
\begin{tabular}{l@{\hspace{0.75\tabcolsep}}l@{\hspace{0.75\tabcolsep}}|l@{\hspace{0.75\tabcolsep}}l|@{\hspace{0.75\tabcolsep}}l@{\hspace{0.75\tabcolsep}}l@{\hspace{0.75\tabcolsep}}l@{\hspace{0.75\tabcolsep}}l@{\hspace{0.75\tabcolsep}}H@{\hspace{0.75\tabcolsep}}l@{\hspace{0.75\tabcolsep}}l@{\hspace{0.75\tabcolsep}}l@{\hspace{0.75\tabcolsep}}l@{\hspace{0.75\tabcolsep}}l@{\hspace{0.75\tabcolsep}}l@{\hspace{0.75\tabcolsep}}l@{\hspace{0.75\tabcolsep}}l@{\hspace{0.75\tabcolsep}}l@{\hspace{0.75\tabcolsep}}l@{\hspace{0.75\tabcolsep}}l@{\hspace{0.75\tabcolsep}}l@{\hspace{0.75\tabcolsep}}H}  
 \toprule  
                                      \textbf{\colorbox{blue!0}{Cluster}}  &
                                      \textbf{\colorbox{blue!0}{Task}}&
                                      \textbf{\colorbox{gray!10}{{B1}}}  &
                                      \textbf{\colorbox{gray!10}{{B2}}}&
                                      \textbf{\colorbox{blue!10}{{M1 }}} &
                                     \textbf{\colorbox{blue!10}{{M2}}}&
                                      \textbf{\colorbox{green!20}{{M3 }}} &
                                                     \textbf{\colorbox{green!20}{{M4   }}} &
                                                     \textbf{\colorbox{green!20}{{M5 }}}  & 
                                                     \textbf{\colorbox{red!10}{{M5 }}} &
                                                     \textbf{\colorbox{red!10}{{M6 }}} &
                                                     \textbf{\colorbox{red!10}{{M7 }}} & 
                                                     \textbf{\colorbox{red!10}{{M8 }}}& 
                                                     \textbf{\colorbox{orange!25}{{M9 }}}  & 
                                                     \textbf{\colorbox{yellow!25}{{M10 }}}  &
                                                     \textbf{\colorbox{yellow!25}{{M11 }}} &
                                                     \textbf{\colorbox{yellow!25}{{M12  }}}&
                                                     \textbf{\colorbox{yellow!25}{{M13  }}} & 
                                                     \textbf{\colorbox{red!35}{{M14 }}} &
                                                     \textbf{\colorbox{blue!25}{{M15 }}} &
                                                     \textbf{\colorbox{purple!35}{{M16 }}}  & 
                                                     \textbf{\colorbox{blue!10}{{M17 }}}  \\
                                                     \toprule 

\multicolumn{1}{l}{\multirow{11}{*}{\textbf{SC}}} 
& abusive\textsuperscript{$\dagger$}  &  $72.68$  &  $71.31$  &  $76.53$  &  $78.36$  &  $75.99$  &  $78.03$  &  $75.61$  &  $75.92$  &  $78.06$  &  $76.22$  &  $76.87$  &  $\bf79.66$  &  $75.49$  &  $76.98$  &  $73.57$  &  $74.43$  &  $77.34$  &  $72.28$  &  $67.68$ \\
& adult\textsuperscript{$\dagger$}  & $89.52$ & $88.49$ & $89.7$ & $90.76$ & $89.67$ & $\bf90.97$ & $89.59$ & $88.97$ & $89.9$ & $89.65$ & $90.18$ & $90.89$ & $90.33$ & $90.09$ & $90.76$ & $88.68$ & $89.35$ & $89.74$ & $88.88$ \\
& age\textsuperscript{$\dagger$}  & $42.68$ & $44.14$ & $44.76$ & $47.11$ & $45.57$ & $46.24$ & $44.93$ & $44.10$ & $44.33$ & $42.02$ & $\bf47.26$ & $46.35$ & $45.89$ & $45.97$ & $45.29$ & $43.29$ & $43.83$ & $45.23$ & $43.61$ \\
&  claim\textsuperscript{$\star$} &  $65.72$  &  $66.66$  &  $70.25$  &  $67.91$  &  $67.38$  &  $67.83$  &  $68.01$  &  $69.74$  &  $69.34$  &  $70.35$  &  $\bf71.53$  &  $69.2$  &  $68.96$  &  $70.32$  &  $65.66$  &  $63.06$  &  $68.81$  &  $66.29$  &  $65.88$ \\
 & dangerous\textsuperscript{$\dagger$}  &  $64.94$  &  $66.31$  &  $\bf67.32$  &  $66.2$  &  $64.96$  &  $67.11$  &  $63.48$  &  $64.72$  &  $62.6$  &  $67.13$  &  $65.66$  &  $66.25$  &  $64.03$  &  $65.31$  &  $66.92$  &  $61.97$  &  $62.83$  &  $64.56$  &  $63.41$ \\
& dialect-B\textsuperscript{$\dagger$} & $84.29$ & $84.78$ & $86.48$ & $86.78$ & $86.92$ & $86.91$ & $94.32$ & $86.64$ & $87.01$ & $87.76$ & $87.21$ & $\bf87.85$ & $86.79$ & $87.40$ & $86.64$ & $84.58$ & $86.57$ & $86.13$ & $85.94$ \\										
&  dialect-R\textsuperscript{$\dagger$}  &  $63.12$  &  $63.51$  &  $67.71$  &  $66.08$  &  $65.21$  &  $66.32$  &  $65.29$  &  $64.63$  &  $67.5$  &  $64.46$  &  $66.34$  &  $66.71$  &  $65.59$  &  $65.05$  &  $68.55$  &  $63.36$  &  $\bf69.22$  &  $63.98$  &  $62.87$ \\
&  dialect-C\textsuperscript{$\dagger$}   &  $25.52$  &  $30.34$  &  $35.26$  &  $35.83$  &  $35.69$  &  $36.06$  &  $36.42$  &  $31.49$  &  $36.33$  &  $27.00$  &  $\bf36.50$  &  $34.36$  &  $33.90$  &  $35.18$  &  $30.83$  &  $27.05$  &  $33.96$  &  $32.99$  &  $28.25$ \\
&  emotion-cls\textsuperscript{$\dagger$}  & $56.79$ & $60.05$ & $63.6$ & $68.85$ & $64.81$ & $\bf70.82$ & $63.21$ & $60.6$ & $64.89$ & $60.98$ & $66.70$ & $68.03$ & $65.25$ & $63.85$ & $64.8$ & $59.66$ & $61.92$ & $62.2$ & $55.22$ \\
& emotion-reg\textsuperscript{$\star$}  & $37.96$ & $52.37$ & $65.37$ & $73.96$ & $67.73$ & $\bf74.27$ & $67.83$ & $62.02$ & $67.64$ & $61.51$ & $70.31$ & $71.91$ & $66.73$ & $65.75$ & $64.34$ & $48.46$ & $66.57$ & $62.77$ & $45.72$ \\
& gender\textsuperscript{$\dagger$}  & $61.78$ & $64.16$ & $64.38$ & $66.65$ & $63.18$ & $\bf67.64$ & $63.51$ & $62.41$ & $64.37$ & $64.24$ & $65.65$ & $66.64$ & $66.38$ & $65.19$ & $64.25$ & $63.37$ & $63.97$ & $64.35$ & $63.50$ \\

 &  hate\textsuperscript{$\dagger$}   &  $72.19$  &  $67.88$  &  $82.41$  &  $81.33$  &  $82.26$  &  $83.54$  &  $80.66$  &  $82.21$  &  $82.39$  &  $81.79$  &  $\bf85.30$  &  $83.88$  &  $81.99$  &  $79.68$  &  $83.38$  &  $74.1$  &  $82.25$  &  $79.77$  &  $74.26$ \\
  &  irony\textsuperscript{$\dagger$}   &  $82.31$  &  $83.13$  &  $83.53$  &  $83.27$  &  $83.83$  &  $83.09$  &  $84.12$  &  $83.63$  &  $84.51$  &  $81.56$  &  $84.62$  &  $\bf85.16$  &  $84.01$  &  $83.07$  &  $81.91$  &  $79.68$  &  $80.91$  &  $83.03$  &  $79.05$ \\
  &  offensive\textsuperscript{$\dagger$}   &  $84.62$  &  $87.18$  &  $89.28$  &  $91.84$  &  $89.55$  &  $\bf92.23$  &  $90.14$  &  $87.5$  &  $90.73$  &  $89.4$  &  $91.89$  &  $91.17$  &  $90.05$  &  $89.32$  &  $90.44$  &  $86.52$  &  $88.76$  &  $87.52$  &  $85.26$ \\

  &  machine G.\textsuperscript{$\star$}  &  $81.4$  &  $84.61$  &  $88.35$  &  $85.14$  &  $87.94$  &  $86.69$  &  $87.69$  &  $87.45$  &  $89.82$  &  $\bf90.66$  &  $87.96$  &  $86.35$  &  $86.73$  &  $88.62$  &  $83.17$  &  $83.35$  &  $87.43$  &  $86.28$  &  $83.91$ \\
  
  &  sarcasm\textsuperscript{$\dagger$}   &  $69.32$  &  $68.42$  &  $73.11$  &  $74.74$  &  $74.16$  &  $76.19$  &  $75.35$  &  $73.46$  &  $74.06$  &  $74.81$  &  $\bf76.83$  &  $75.82$  &  $74.17$  &  $75.18$  &  $72.57$  &  $69.94$  &  $72.02$  &  $73.11$  &  $71.92$ \\

& sentiment\textsuperscript{$\dagger$}  & $78.99$ & $77.21$ & $77.78$ & $79.08$ & $78.60$ & $80.83$ & $78.85$ & $78.45$ & $80.50$ & $79.56$ & $\bf80.86$ & $80.33$ & $79.51$ & $79.75$ & $77.8$ & $76.76$ & $78.68$ & $78.46$ & $76.46$ \\ 
\hline

\multicolumn{1}{l}{\multirow{3}{*}{\textbf{SP}}} 
 &  ner-anerc.\textsuperscript{$\star$}   &  $85.92$  &  $86.76$  &  $90.27$  &  $86.59$  &  $90.83$  &  $87.86$  &  $90.87$  &  $90.68$  &  $\bf90.85$  &  $90.17$  &  $90.03$  &  $87.5$  &  $89.27$  &  $90.71$  &  $83.61$  &  $82.94$  &  $89.52$  &  $88.77$  &  $86.54$ \\
  &  ner-aqmar\textsuperscript{$\star$}   &  $75.95$  &  $76.16$  &  $80.72$  &  $74.57$  &  $\bf81.70$  &  $74.22$  &  $80.19$  &  $77.34$  &  $79.2$  &  $73.43$  &  $77.66$  &  $73.72$  &  $76.84$  &  $78.54$  &  $73.77$  &  $70.71$  &  $74.97$  &  $79.5$  &  $73.15$ \\
  & pos-dia\textsuperscript{$\dagger$} & $92.04$ & $92.78$  &  $92.92$  &  $94.14$  &$93.92$  & $93.38$  &   xx  &  $91.65$  &  $93.79$  &  $94.70$   & $93.554$ & $\bf94.70$ & $93.95$ & $94.37$ & $93.95$ & $92.05$ & $92.05$ & $93.24$ & $92.57$ &  $xx$ \\ 
   &   pos-xglue\textsuperscript{$\star$}    &   $57.68$ & $69.37$ & $51.39$ & $55.02$ & $52.55$ & $55.45$ & $34.65$&  $34.65$ & $37.84$ & $41.28$ & $26.61$ & $41.36$ & $27.70$ & $32.89$ & $10.37$ & $17.04$ & $62.58$ & $\bf63.89$  & $42.40$ & $32.10$  \\ \hline
  
 \multicolumn{1}{l}{\multirow{3}{*}{\textbf{NLI}}} 

 &  ans-st\textsuperscript{$\star$}   &  $84.49$  &  $81.00$  &  $91.77$  &  $73.82$  &  $91.02$  &  $80.57$  &  $88.87$  &  $87.59$  &  $\bf93.23$  &  $92.33$  &  $90.17$  &  $50.2$  &  $82.49$  &  $89.21$  &  $46.01$  &  $71.81$  &  $85.31$  &  $82.86$  &  $80.02$ \\
 &  baly-st\textsuperscript{$\star$}   &  $34.48$  &  $38.27$  &  $45.63$  &  $29.07$  &  $49.34$  &  $36.52$  &  $49.82$  &  $\bf51.19$  &  $46.63$  &  $41.32$  &  $37.12$  &  $31.58$  &  $48.94$  &  $49.67$  &  $30.58$  &  $48.85$  &  $49.21$  &  $49.22$  &  $47.19$ \\ 
  &  xlni\textsuperscript{$\star$}   &  $61.88$  &  $65.06$  &  $67.22$  &  $60.50$  &  $68.17$  &  $62.22$  &  $67.95$  &  $64.69$  &  $67.93$  &  $\bf70.20$  &  $66.67$  &  $55.67$  &  $63.82$  &  $66.02$  &  $54.29$  &  $61.18$  &  $66.53$  &  $62.15$  &  $61.62$ \\\hline 
  
\multirow{2}{*}{\textbf{STS}}  

  &  sts-r\textsuperscript{$\star$} \textsuperscript{$\ddagger$}  &  $63.91$  &  $62.24$  &  $73.00$  &  $63.48$  &  $71.90$  &  $66.12$  &  $73.19$  &  $71.27$  &  $75.4$  &  $\bf76.01$  &  $70.50$  &  $41.15$  &  $70.61$  &  $74.42$  &  $71.23$  &  $70.13$  &  $73.68$  &  $73.56$  &  $66.75$ \\
  &  sts-c\textsuperscript{$\star$} &  $62.34$  &  $63.35$  &  $85.95$  &  $74.43$  &  $96.73$  &  $63.47$  &  $85.98$  &  $96.81$  &  $64.11$  &  $64.24$  &  $63.52$  &  $84.11$  &  $63.28$  &  $\bf97.10$  &  $59.57$  &  $96.41$  &  $85.87$  &  $96.69$  &  $62.91$ \\\hline 
  \textbf{TC} &  
  topic\textsuperscript{$\star$}   &  $92.55$  &  $93.53$  &  $94.17$  &  $93.53$  &  $93.96$  &  $93.9$  &  $94.32$  &  $94.31$  &  $\bf94.58$  &  $94.11$  &  $94.02$  &  $93.32$  &  $93.72$  &  $94.38$  &  $93.18$  &  $93.41$  &  $94.05$  &  $93.86$  &  $93.27$ \\ \hline

 \multirow{1}{*}{\textbf{QA}}  
   &  arlue-qa\textsuperscript{$\star$}  &  $56.39$  &  $56.51$  &  $57.65$  &  $49.35$  &  $61.5$  &  $57.9$  &  $61.46$  &  $56.79$  &  $\bf61.56$  &  $60.70$  &  $57.65$  &  $45.27$  &  $53.98$  &  $57.46$  &  $30.91$  &  $52.11$  &  $58.71$  &  $55.94$  &  $53.89$ \\ \hline
 
\textbf{WSD} &
  ar-wsd\textsuperscript{$\star$}   &  $69.82$ & $52.90$& $33.29$ & $72.94$ & $71.01$ & $33.28$   &  $xx$  &  $51.72$  &  $\bf76.68$  &   $73.54$  & $72.92$  &  $70.13$  &  $74.12$  &  $75.86$   & $65.18$  &  $75.68$  &  $74.31$  &  $69.76$  &  $47.19$  &  $xx$  \\  \hline

  \colorbox{red!10}{\textbf{Avg. DIA\textsuperscript{$\dagger$}  }}& & $69.39$ & $69.98$& $72.98$ & $74.07$  & $72.95$ &\colorbox{red!10}{$\bf74.62$} &  xx & $71.76$ & $73.40$ & \underline{\bf$74.47$} & $71.97$ & \underline{\bf$74.47$}
  & $73.18$ &$73.06$   & $72.79$ & $69.70$&  $72.25$ & $71.65$& $69.10$ & $xx$  \\   \hline 
 \colorbox{green!15}{\textbf{Avg. MSA\textsuperscript{$\star$}  }} &  &$66.75$ & $67.77$ & $73.91$ &$65.76$ & \colorbox{green!15}{$\bf75.13$} & $67.17$ & $xx$ & $71.16$& $72.48$ & $70.65$ & $70.37$ 
 & $64.39$ & $69.09$ & $73.64$ & \underline{$59.42$} & $66.80$ & $74.20$ & $72.15$ &$47.19$& $xx$ \\ \hline 
 							
  \toprule  
  
\multicolumn{2}{l}{\colorbox{cyan!25}{\textbf{\ourbench\textsubscript{$score$}}}}&  $68.07$ & $68.88$ & $73.45$ &$69.91$ & \colorbox{cyan!25}{$\bf74.04$} & $72.95$ &$xx$ & $71.46$ & $72.94$ & $72.56$ 
& $71.17$ & $69.43$ & $71.13$ & \underline{$73.35$}  & $66.10$ & $69.15$ & $73.23$ & $71.90$ & $58.14$ & $xx$\\ 			
\toprule  

\end{tabular} }

\caption{\small{Performance of  Arabic Bert-based  models on \ourbench~Test  splits (F$_1$) $^{\ddagger}$~Metric for STSP taks is spearman correlation.  \colorbox{gray!15}{\textbf{B1, B2}:} Two baselines mBERT~\cite{devlin2019bert} and XLM-R~\cite{liu2019roberta}. \colorbox{blue!10}{\textbf{M1, M2}:}  ARBERT, MARBERT~\cite{abdul2020arbert}. \colorbox{green!20}{\textbf{M3, M4}:} ARBERT\textsubscript{V2} and MARBERT\textsubscript{V2}.  \colorbox{red!10}{\textbf{M5, M6, M7, and M8}:}  AraBERT\textsubscript{v1[v2, tw]}, and AraElectra~\cite{antoun2020arabert, antoun2020araelectra}.  \colorbox{orange!25}{\textbf{M9}:} QARiB~\cite{chowdhury2020} \colorbox{yellow!25}{\textbf{M10, M11, M12, and M13}:} CamelBERT\textsubscript{mix[msa, da, ca]}~\cite{CAMeLBERT2021}.  \colorbox{red!35}{\textbf{M14}:} GigaBERT\textsubscript{v4}~\cite{chowdhury2020}.  \colorbox{blue!25}{\textbf{M15}:} Arabic BERT~\cite{chowdhury2020}. \colorbox{purple!25}{\textbf{M16}}: Arabic Albert~\cite{lan2020gigabert}. \textbf{Avg. DIA and Avg. MSA:}~The average of dialect and MSA tasks. \colorbox{cyan!25}{\textbf{\ourbench\textsubscript{$score$}}}: Average overall Dia and MSA tasks. \textsuperscript{$\star$}MSA tasks. \textsuperscript{$\ddagger$}DIA tasks.   A task is  considered  as an  MSA if it has more than $98$\% samples  predicted as MSA using an MSA vs. DIA classifier (see Table~\ref{tab:SuperARLUE_msa_dia}).}}\label{tab:res_ar_LMs}

\end{table*} 

 \section{Language  Models}\label{sec:LM}
In this section, we list multilingual PLMs that include Arabic in its coverage only by name, for space but provide a description of each of them in Appendix~\ref{appdx_sec:LM}.
  
\noindent\textbf{Multilingual LMs}. These are mBERT~\cite{devlin2019bert} , XLM-R~\cite{conneau-etal-2020-unsupervised}, GigaBERT~\cite{lan2020gigabert}, and mT5~\cite{xue2020mt5}. 

\noindent\textbf{Arabic LMs}. These are AraBERT~\cite{antoun2020arabert}, ArabicBERT~\cite{safaya2020}, Arabic-ALBERT~\cite{ali_safaya_2020_4718724}, QARiB~\newcite{chowdhury2020}, ARBERT \& MARBERT~\cite{abdul2020arbert}, CamelBERT~\cite{CAMeLBERT2021}, JABER and SABER~\cite{ghaddar2021jaber}, and AraT5~\cite{nagoudi2021arat5}. 

\noindent Table~\ref{appdx_tab:models_config_comp} (Appendix~\ref{appdx_sec:LM}) shows a comparison between the multilingual as well as the Arabic PLMs in terms of (1) training data size, (2) vocabulary size, (3) language varieties, and (4) model configuration and architecture.

\section{Model Evaluation on~\ourbench}\label{sec:Eval}

This section shows experimental settings and performance of  $18$ multilingual and Arabic language models on \ourbench~downstream tasks.\footnote{We exclude JABER and SABER~\cite{ghaddar2021jaber} from the evaluation  as these are not supported by the Transformers library.}

\noindent\textbf{Baselines.}
For comparison, we finetune the multilingual language models  mBERT and XLM-R\textsubscript{Base}  on  all training data of \ourbench~benchmark. 

\noindent\textbf{Evaluation.} For all models and baselines, across all tasks, we identify the best model on the respective development data split (Dev) and blind-test on the testing split (Test).  We methodically evaluate each task cluster, ultimately reporting a single \textit{\ourbench~score} following~\newcite{wang2018glue,abdul2020arbert}.~\ourbench~score is simply the macro-average of the different scores across all tasks and task clusters, where  each task is weighted equally. We compute the~\ourbench~score for all $18$ language models.
 
\noindent\textbf{Results.} We present results of all language models and the baselines on each task cluster of \ourbench~independently using the relevant metric, for both Dev (see Table~\ref{tab:res_ar_LMs_dev} in Appendix~\ref{appx:eval}) and Test (see Table~\ref{tab:res_ar_LMs}). As Table~\ref{tab:res_ar_LMs} shows, ARBERT\textsubscript{v2} (M3) achieves the highest \ourbench~score  across all the tasks and also for MSA tasks only\footnote{We consider a task an MSA task if it has more than 98\% of samples  predicted as MSA using the MSA vs. DA classifier (see Table~\ref{tab:SuperARLUE_msa_dia}).} (\textit{with \ourbench~score=}$74.04$ and \textit{Avg. MSA score=}$75.13$), followed by CamelBERT\textsubscript{msa} (M11) in both cases with $73.35$ and $73.64$, respectively. Regarding the dialect tasks, we note that MARBERT\textsubscript{v2} (M4) achieves the best dialect ORCA score (\textit{Avg. DA score=}$74.62$) followed by QARIB (M9) with $74.47$. 
\noindent We also note that  AraELECTRA (M8) achieves the best results  in six individual tasks out of $26$, followed by MARBERT\textsubscript{v2} (M4) which excels in five individual tasks.

\noindent\textbf{Analysis.}
As our experiments imply, \ourbench~ allow us to derive unique insights. Example insights that can be derived from Table~\ref{tab:res_ar_LMs} are: (a) a model such as M6 (i.e., AraBERTv2) that is pretrained with historical data (AlSafeer newspaper) would excel on old datasets (e.g., TC, QA, and WSD); while M4 (i.e., MARBERTv2) excels especially on datasets from social media since it is pretrained with a large Twitter collection. In addition, since ORCA arranges the metrics into one dedicated to dialect, another to MSA, and a third to both (ORCA score), it is much easier to compare model performance across the DA-MSA dimensions.


\section{Analysis of Model Computational Cost}\label{sec:diss}

We also compare the Arabic language models in terms of computational cost using the average time needed for convergence (in minutes) and average number of epochs to convergence as identified on Dev sets. For this, we finetune all models for a maximum of $25$ epochs on all \ourbench~tasks. \textit{We report results in terms of average of three runs.} Figure~\ref{fig:25_epoch} (Appendix~\ref{appx_sec_model_analysis}) shows for each model the \textit{total time needed for convergence} (out of $25$ epochs), and Figure~\ref{fig:avg_conv} (Appendix~\ref{appx_sec_model_analysis}) shows \textit{average convergence time} and \textit{average number of epochs till convergence}. As ~\ref{fig:25_epoch} (Appendix~\ref{appx_sec_model_analysis}) shows, Arabic Albert is  the fastest model ($52.26$ min) to finetune for $25$ epochs, but it achieves the lowest \ourbench~score. Excluding Arabic Albert, we observe a near constant time (between $60.32$-$63.69$ mins) for all other models. Among the top five models, as Figure~~\ref{fig:avg_conv} (Appendix~\ref{appx_sec_model_analysis}) shows, we also observe that ARBERT\textsubscript{v1} is the fastest (in terms of average convergence time and number of epochs needed to converge) and is followed by QARiB. 

\section{Conclusion}\label{sec:conc}

We  presented \ourbench, a large and diverse benchmark for Arabic natural language understating tasks composed of $60$ datasets that are arranged in seven task clusters. To facilitate future research and adoption of our benchmark, we offer a publicly-available interactive leaderboard with a useful suite of tools and extensive meta-data. In addition, we provide a comprehensive and methodical evaluation as well as meaningful comparisons between $18$ multilingual and Arabic language models on~\ourbench. We also compare the models in terms of computing needs. As our results show,~\ourbench~is challenging and we hope it will help standardize comparisons and accelerate progress both for Arabic and multilingual NLP.


\section{Limitations}\label{sec:limit}
We identify the following limitations:

\begin{enumerate}

\item Although we strive to include tasks in all Arabic varieties, available downstream datasets from certain countries such as Mauritania and Djibouti are almost nonexistent and so are not covered in~\ourbench. In addition, there is a need in the community to create more datasets for several Arabic dialects. This includes, for example, dialects such as Iraqi, Sudanese, and Yemeni. With the introduction of more datasets for such dialects,~\ourbench's coverage can be further extended. Regardless, as Figure~\ref{fig:map_countery} (Appendix~\ref{appx_sec_model_data}) shows,  \ourbench~datasets are quite diverse from a geographical perspective.

\item Although~\ourbench~currently covers both dialectal Arabic (DA) and MSA, it does not pay as much attention to the classical variety of Arabic (CA) due to historical reasons. That is, the community did not invest as much efforts creating and releasing datasets involving CA. However, as more unlabeled datasets become available and with an undergoing positive change in the culture around data sharing, this is likely to change in the near future. Again, this will make it possible to extend~\ourbench~to better cover CA in the future.

\item Although benchmarks in general are useful in encouraging standardize evaluations and meaningful comparisons, and can help motivate progress within the community, they also run the risk of contributing to a culture of leaderboard chasing that is not necessarily useful. That is, although scientific research advances due to competition, it also thrives through partnerships and collaborations that bring the best from diverse groups. It is in the context of this collaborative culture that we hope~\ourbench~will be perceived and used.

\end{enumerate}
\section{Ethics Statement and Broad Impact}\label{sec:ethics}
\textbf{Encouraging standardized evaluations and contributing to a collaborative research culture.} Similar to some other research communities, progress in the Arabic NLP community has been hampered for a long time by absence of standardized and meaningful evaluations for some tasks. This is due to several reasons, including the culture around data sharing but also as a result of insufficient funding and lack of strong training programs. This has made it challenging to measure progress. The Arabic NLP community is now expanding, and a culture of collaboration is being built as part of the larger positive developments within the overall NLP community itself. As such, it is now ripe time to introduce benchmarks that can help this ongoing progress. We hope there will be wide adoption of~\ourbench~and that our work will trigger more efforts to create more benchmarks, including for newer tasks in what could be a virtuous cycle.

\noindent\textbf{Data privacy.} Regarding data involved in~\ourbench, we develop the benchmark using data from the public domain. For this reason, we do not have serious concerns about privacy. 

\noindent\textbf{Sufficient assignment of credit to individual data sources.} Another important consideration in benchmarking is how credit is assigned to creators of the individual datasets. To ensure sufficient credit assignment, we refer users to the original publications, websites, GitHub repositories where a dataset originated and link all these sources in our leaderboard. We also provide bibliographic entries for all these sources that users can easily copy and paste in order to cite these original sources. By encouraging citation of original sources in any publications in the context of~\ourbench~use, we hope to afford additional visibility to many of the individual datasets. 





 \section*{Acknowledgements}\label{sec:acknow}
MAM gratefully acknowledges support from Canada Research Chairs (CRC), the Natural Sciences and Engineering Research Council of Canada (NSERC; RGPIN-2018-04267), the Social Sciences and Humanities Research Council of Canada (SSHRC; 435-2018-0576; 895-2020-1004; 895-2021-1008), Canadian Foundation for Innovation (CFI; 37771), Digital Research Alliance of Canada (DRAG),\footnote{\href{https://alliancecan.ca}{https://alliancecan.ca}} UBC ARC-Sockeye,\footnote{\href{https://arc.ubc.ca/ubc-arc-sockeye}{https://arc.ubc.ca/ubc-arc-sockeye}} Advanced Micro Devices, Inc. (AMD), and Google. Any opinions, conclusions or recommendations expressed in this material are those of the author(s) and do not necessarily reflect the views of CRC, NSERC, SSHRC, CFI, DRAG, AMD, Google, or UBC ARC-Sockeye.
\normalem
\bibliography{tacl2022}
\bibliographystyle{acl_natbib}

\appendix
 
\clearpage
\appendixpage
\addappheadtotoc
\numberwithin{figure}{section}
\numberwithin{table}{section}

In this appendices, we provide an addition which organized as follows:\\

\noindent \textbf{Sections list}:
\begin{itemize}\setlength\itemsep{.2em}
    \item Language Models. (Section~\ref{appdx_sec:LM})
    \begin{itemize}
        \item Multilingual LMs. (Subsection~\ref{appdx_sec:Mult_LM})
        \item Arabic LMs. (Subsection~\ref{appdx_sec:Arabic_LM})
     \end{itemize}
     \item X-Specific Benchmarks. (Section~\ref{appdx_sec:xs_benchmark})
     \item \ourbench~Evaluation. (Section~\ref{appx:eval})
     \item Public leaderboard. (Section~\ref{apdx:leaderboard}
     \item Analysis of Model Computational Cost. (Section~\ref{appx_sec_model_analysis})
     \item \ourbench~Data. (Section~\ref{appx_sec_model_data})
\end{itemize}

\noindent \textbf{Tables and Figures List}:

\begin{itemize}
    \item Configuration comparisons of Arabic PLMs and multilingual PLMs  (Table~\ref{appdx_tab:models_config_comp}).
    \item Performances of Arabic BERT-based models on \ourbench~Dev splits. (Table~\ref{tab:res_ar_LMs_dev})
    \item Randomly picked examples from the dialectal portion of~\ourbench~Train datasets. (Table~\ref{tab:samples} )
    
    \item Models' ORCA scores across all $29$ tasks in~\ourbench~benchmark. (Figure~\ref{appdx_fig:preformace_all_tasks}
    \item Models' ORCA scores across  all clusters in~\ourbench~benchmark. (Figure~\ref{appdx_fig:preformace_all_clusters})
    \item Models' F\textsubscript{1} scores across all tasks in the sentence classification cluster. (Figure~\ref{appdx_fig:preformace_SC_clusters})
    \item An example of tasks sorted alphabetically. (Figure~\ref{fig:screen_onetask})
    \item Detailed scores by all models for a given task. (Figure~\ref{fig:screen_tasks_leaderboard})
    \item Detailed information about each task cluster and associated tasks, with each task assigned an identifier, language variety, evaluation metric, a link to the dataset website/GitHub/paper and  bibliographic information. (Figure~\ref{fig:screen_tasks_leaderboard})
    \item The average number of epochs (in orange), and time needed to converge (mins, in blue) for all the studied PLMs across all~\ourbench~ tasks. (Figure~\ref{fig:avg_conv})
    \item The time needed in minutes to finetune ($25$ epochs). We compute the average time of three runs across all~\ourbench~tasks. (Figure~\ref{fig:25_epoch})
    \item The predicted country-level distribution, in percentage, in the dialectal portion of~\ourbench. (Figure~\ref{fig:map_countery})
    
\end{itemize}

 \section{Language  Models}\label{appdx_sec:LM}
\begin{table*}[t]
 \renewcommand{\arraystretch}{1.12}
\resizebox{1\textwidth}{!}{%
\begin{tabular}{lllrrcrHHc}
\toprule
& \multicolumn{1}{c}{\multirow{2}{*}{\textbf{Models}}} & \multicolumn{3}{c}{\textbf{Training Data}}         & \multicolumn{3}{c}{\textbf{Vocabulary}}            & \multicolumn{2}{c}{\textbf{Configuration}} \\  \cline{3-9} 
& \multicolumn{1}{c}{}      &    \multicolumn{1}{l}{\textbf{Type}} &       \multicolumn{1}{l}{\textbf{Text Size (ar)}}             & \multicolumn{1}{c}{\textbf{Tokens} (ar/all)} & \multicolumn{1}{c}{\textbf{Tok.}} & \multicolumn{1}{c}{\textbf{Size}} & \textbf{Cased} & \textbf{Arch.}              & \textbf{\#Param.}            \\ \toprule

\multirow{1}{*}{\rotatebox[origin=]{90}{\textbf{\colorbox{blue!10}{\small{ML LMs}}}}} & mBERT       & MSA  &  $1.4$GB            & $153$M/$1.5$B   & WP       & $110$K   & yes   & base              & $110$M    \\ 
&XLM-R  & MSA  &  $5.4$GB      & $2.9$B/$295$B & SP   & $250$K  & yes   & base              & $270$M    \\ 
& GigaBERT & MSA  &  $42.4$GB          &  $4.3$B/$10.4$B & WP       & 50k  & no    & base              & 125M  \\

\hdashline 

& ARBERT   & MSA & 61GB & $6.2$B & WP       & $100$K      & no    & base              & $163$M    \\ 
&\colorbox{green!15}{ARBERT\textsubscript{v2}}  & MSA, DA   & $243$GB  & $27.8$B  & WP       & $100$K      & no    & base              & $163$M    \\ 
& MARBERT  & MSA, DA   & 128GB  & $15.6$B  & WP       & $100$K      & no    & base              & $163$M    \\ 
& MARBERT\textsubscript{v2}   & MSA & $198$GB & $21.4$B & WP       & $100$K      & no    & base              & $163$M    \\

\multirow{8}{*}{\rotatebox[origin=c]{90}{\textbf{\colorbox{blue!10}{Arabic LMs}}}} & AraBERT     & MSA & $27$GB & $2.5$B & WP   & $64$K   & no    & base              & $135$M    \\   

& AraELECTRA     & MSA  & $77$GB  & $8.8$B & WP   & $64$K   & no    & base              & $135$M    \\ 
& ArabicBERT     &  MSA  & $95$GB & $8.2$B & WP   & $64$K   & no    & base              & $135$M    \\ 
& Arabic-ALBERT     & MSA & $33$GB & $4.4$B & WP   & $32$K   & no    & base              & $110$M    \\ 
& QARiB     &    MSA, DA & $97$GB & $14$B & WP   & $64$K   & no    & base              & $135$M    \\ 
& CAMeLBERT     &   MSA, DA, CA & $167$GB & $8.8$B & WP   &$30$K   & no    & base              & $108$M    \\ 
& JABER     &   MSA & $115$GB & $-$ &  BBPE  &  $64$K   & no    & base              & $135$M    \\ 
& SABER     &   MSA& $115$GB & $-$ &  BBPE &   $64$K   & no    & base              & $135$M    \\ 
& AraT5     &   MSA, DA  & $248$GB & $29$B & SP   &$110$K   & no    & base              & $220$M    \\

\toprule
\end{tabular}%
}

\caption{Configuration comparisons of  Arabic pre-trained LMs and multilingual  LMs  which covered Arabic.\textbf{ WP}: WordPiece~\cite{schuster2012japanese}. \textbf{SP}: SentencePiece~\cite{kudo2018sentencepiece}. \textbf{BBPE}: Byte-level Byte Pair Encoding~\cite{wei2021training}.  \textbf{\colorbox{green!15}{ARBERT\textsubscript{v2}:} }a new model proposed in this paper.   }
\label{appdx_tab:models_config_comp}
\end{table*}

In this section, we provide a description of the multilingual MLM that include Arabic in its coverage. 
\subsection{Multilingual LMs} \label{appdx_sec:Mult_LM}
\noindent\textbf{mBERT} is the multilingual version of BERT~\cite{devlin2019bert} which is
 a multi-layer bidirectional encoder representations from Transformers~\cite{vaswani2017attention} trained with a masked language modeling.~\newcite{devlin2019bert} present two architectures: \textit{Base} and \textit{Large}.  BERT models were trained on  English Wikipedia\footnote{\href{https://www.wikipedia.org/}{https://www.wikipedia.org/}} and  BooksCorpus~\cite{Zhu_2015_ICCV}. mBERT  is trained on Wikipedia for $104$ languages (including $\sim153$M Arabic tokens). \\
\textbf{XLM-R}~\cite{conneau-etal-2020-unsupervised} is a transformer-based multilingual masked language model pre-trained on  more than $2$TB  of filtered CommonCrawl data in $100$ languages, including Arabic ($2.9$B tokens).  XLM-R uses a Transformer model trained a multilingual version of masked language modeling of XLM~\cite{lample2019cross}. XLM-R  comes with  two sizes and architectures: \textit{Base} and  \textit{Large}. The XLM-R\textsubscript{Base}  architecture contains $12$ layers, $12$ attention heads, $768$ hidden units,  and  $270$M parameters. The XLM-R\textsubscript{Large}  architecture has   $124$ layers, $16$ attention heads, $1024$ hidden units,  and  $550$M parameters. While both XLM-R models use the same masking objective as BERT, they do not include the next sentence prediction objective used in BERT.   

\noindent\textbf{GigaBERT}~\cite{lan2020gigabert}  is a customized bilingual BERT-based model for Arabic and English  pretrained on a corpus of ~$10$B tokens collected from different sources, including:  English and Arabic Gigaword corpora,\footnote{\href{https://catalog.ldc.upenn.edu/LDC2011T07}{https://catalog.ldc.upenn.edu/LDC2011T07}}, OSCAR~\cite{suarez2019asynchronous}, and  Wikipedia. GigaBERT is designed specifically for  zero-shot transfer learning from English to Arabic  on information extraction  tasks. 

\noindent\textbf{mT5}~\cite{xue2020mt5} is the multilingual version of \textbf{T}ext-\textbf{t}o-\textbf{T}ext \textbf{T}ransfer \textbf{T}ransformer model (T5)   \cite{raffel2019exploring}.  The T5 model architecture is essentially an encoder-decoder Transformer similar in configuration and size to  BERT\textsubscript{Base}. The T5 model treats every  text-based language task as a ``text-to-text" problem, (i.e. taking text format as input and producing new text format as output), where multi-task learning  is applied with several  NLP tasks: question answering, document summarization, machine translation, and sentiment classification.  mT5  is trained on the ``Multilingual Colossal Clean Crawled Corpus" (or mC4 for short), which is  $\sim26.76$TB for $101$ languages (including Arabic with more than $\sim57$B tokens) generated from $71$ Common Crawl dumps.
 \subsection{Arabic LMs}\label{appdx_sec:Arabic_LM}
Several Arabic LMs have been developed. We describe the most notable among these here. \\


\noindent\textbf{AraBERT}~\cite{antoun2020arabert} is the first  pretrained language model proposed for Arabic. It is based on  the two  BERT\textsubscript{Base} and  BERT\textsubscript{Large} architectures. AraBERT\textsubscript{Base} ~\cite{antoun2020arabert}  is trained on $24$GB of Arabic text ($70$M sentences and $3$B tokens) collected from  Arabic Wikipedia, Arabic news,  Open Source International dataset (OSIAN)~\cite{zeroual2019osian}, and $1.5$B words corpus from~\cite{elkhair-2016}.  
In order to train BERT\textsubscript{Large}~\newcite{antoun2020araelectra} use the same AraBERT\textsubscript{\textit{Base}} data augmented with the unshuffled Arabic OSCAR dataset~\cite{suarez2019asynchronous} and news articles provided by As-Safir newspaper\footnote{\href{https://www.assafir.com/}{https://www.assafir.com/}} ($77$GB or $8.8$B tokens) . The augmented data is also used to train AraELECTRA\textsubscript{\textit{Large}}--an Arabic language model that employs an ELECTRA objective~\cite{clark2020electra}.

\noindent\textbf{ArabicBERT} is an Arabic BERT-based model proposed by~\newcite{safaya2020} Authors pretrain fourr variants: ArabicBERT\textsubscript{Mini}, ArabicBERT\textsubscript{Medium}, ArabicBERT\textsubscript{Base}, ArabicBERT\textsubscript{Large}.\footnote{\href{https://github.com/alisafaya/Arabic-BERT}{https://github.com/alisafaya/Arabic-BERT}} The models are pretrained on unshuffled Arabic OSCAR~\cite{suarez2019asynchronous}, Arabic Wikipedia, and other Arabic resources which sum up to ~95GB of text ($\sim 8.2$B tokens).

\noindent\textbf{Arabic-ALBERT}~\cite{ali_safaya_2020_4718724} is an Arabic language representation  model based on \textbf{A} \textbf{L}ite \textbf{Bert} (ALBERT)~\cite{lan2019albert}.  ALBERT is a Transformer-based neural network architecture (similar to BERT and XLM-R)  with two parameter reduction techniques proposed to increase the training speed and lower memory consumption of the BERT model. Arabic-ALBERT is pretrained on $\sim4.4$B tokens extracted from Arabic OSCAR~\cite{suarez2019asynchronous} and Arabic Wikipedia. Arabic-ALBERT comes with  three different architectures:  Arabic-ALBERT\textsubscript{Base}, Arabic-ALBERT\textsubscript{Large}, Arabic-ALBERT\textsubscript{XLarge}. 

\noindent\textbf{QARiB.}~\newcite{chowdhury2020} propose  the \textbf{Q}CRI \textbf{AR}abic and D\textbf{i}alectal \textbf{B}ERT   (QARiB) model.  QARiB is trained on a collection of $97$GB of Arabic Text (14B tokens) on both MSA  ($180$ Million sentences) and Twitter data ($420$ Million tweets). Authors use  the Twitter API to collect Arabic tweets, keeping only tweets identified as Arabic by Twitter language filter. For MSA data in QARiB is a combination of Arabic Gigaword,\footnote{\href{https://catalog.ldc.upenn.edu/LDC2011T11}{https://catalog.ldc.upenn.edu/LDC2011T11}}, Abulkhair Arabic Corpus~\cite{elkhair-2016}, and OPUS~\cite{tiedemann2012parallel}.

\noindent\textbf{ARBERT}~\cite{abdul2020arbert}  is a pretrained language model focused on MSA. ARBERT is trained using the same architecture as BERT\textsubscript{Base} with a vocabulary of $100$K WordPieces, making $\sim163$M parameters. ARBERT exploits a  collection of Arabic datasets comprising $61$GB of text  ($6.2$B tokens) from the following sources:  El-Khair~\newcite{elkhair-2016}, Arabic Gigaword,\footnote{ \href{https://catalog.ldc.upenn.edu/LDC2009T30}{https://catalog.ldc.upenn.edu/LDC2009T30}}, OSCAR~\cite{suarez2019asynchronous}, OSIAN~\cite{zeroual2019osian},  Arabic Wikipedia, and Hindawi Books.\footnote{\href{https://www.hindawi.org/books/}{https://www.hindawi.org/books}}  

\noindent\textbf{ARBERT\textsubscript{v2}.} We provide a new Arabic version of \noindent\textbf{ARBERT}, by further pretraining ARBERT on $243$GB MSA dataset ($70$GB MSA data from various sources and $173$GB extracted and cleaned from the Arabic part of the multilingual Colossal Clean Crawled Corpus (mC4)~\cite{xue2020mt5}.

\noindent\textbf{MARBERT}~\cite{abdul2020arbert} is a pretrained language model focused on both dialectal Arabic  and MSA. This model is trained on a sample of $1$B Arabic tweets ($128$GB of text, ~$15.6$B tokens). In this dataset,  authors keep only tweets with at least $3$ Arabic words (based on character string matching) regardless of whether the tweet has non-Arabic string or not. MARBERT uses the same vocabulary size ($100$K WordPieces) and network architecture as ARBERT (BERT\textsubscript{Base}), but without the next sentence prediction objective since tweets are short.
\noindent\textbf{MARBERT\textsubscript{v2}}.~\newcite{abdul2020arbert} further pretrain MARBERT with additional data using a larger  sequence length of $512$ tokens for $40$ epochs.


\noindent\textbf{CamelBERT}~\cite{CAMeLBERT2021} is pre-trained using BERT\textsubscript{Base} architecture on four types of Arabic datasets:  MSA ($107$GB),  dialectal Arabic ($54$GB), classical Arabic ($6$GB), and a mixture of the last three datasets ($167$GB). CamelBERT is trained using  a small  vocabulary  of $30$K tokens (in WordPieces).

\noindent\textbf{JABER and SABER}~\cite{ghaddar2021jaber} are BERT-based models  (Base and Large)  pretraind on $115$GB  of text data collected from Common Crawl (CC), OSCAR~\cite{suarez2019asynchronous}, OSIAN~\cite{zeroual2019osian}, El-Khair~\newcite{elkhair-2016}, and Arabic Wikipedia. In order to overcome the out-of-vocabulary problem and improve the representations of rare words, JABER is trained using a Byte-level Byte Pair Encoding (BBPE)~\cite{wei2021training} tokenizer with  a vocabulary size of $64$K.

\noindent\textbf{AraT5}~\cite{nagoudi2021arat5} is an  Arabic text-to-text Transformer  model dedicated to MSA and dialects. It is essentially an encoder-decoder Transformer similar in configuration and size to T5~\cite{raffel2019exploring}. AraT5 is trained on more than $248$GB of Arabic text ($70$GB MSA and $178$GB tweets), where the data is from the following sources: AraNews~\cite{nagoudi2020machine}, El-Khair~\newcite{elkhair-2016}, Gigaword,\footnote{ \href{https://catalog.ldc.upenn.edu/LDC2009T30}{https://catalog.ldc.upenn.edu/LDC2009T30}}, OSCAR~\cite{suarez2019asynchronous}, OSIAN~\cite{zeroual2019osian},  Wikipedia Arabic, and Hindawi Books.\footnote{\href{https://www.hindawi.org/books/}{https://www.hindawi.org/books}}

Table~\ref{appdx_tab:models_config_comp}  shows a comparison between the multilingual as well as the Arabic language models in terms of (1) training data size,  (2) vocabulary size, (3)  language varieties, and (4) model configuration and architecture.

\section{X-Specific Benchmarks}\label{appdx_sec:xs_benchmark}
\noindent\textbf{CLUE.}~\newcite{xu2020clue} introduce CLUE, a benchmark for Chinese NLU. It covers nine tasks spanning single-sentence/sentence-pair classification, text classification,  coreference resolution,  semantic similarity, and question answering.

\noindent\textbf{FLUE.}~\newcite{le2020flaubert} offer FLUE, a French NLU benchmark involving six datasets with different levels of difficulty, degrees of formality, and  domains. FLUE is arranged into three tasks: text classification, paraphrasing, and NLI.

\noindent\textbf{IndoNLU.}~\newcite{wilie2020indonlu} present IndoNLU, a benchmark for Bahasa Indonesian NLU with $12$~downstream tasks organized into five task clusters: sentence classification, structure protection, text classification, semantic similarity, and question answering. 

\noindent\textbf{JGLUE.}~\newcite{kurihara2022jglue} propose JGLUE, a Japanese NLU benchmark consisting of six datasets arranged into three task clusters: sentence classification, text classification, and question answering.

\noindent\textbf{KorNLI and KorSTS.}~\newcite{ham2020kornli} release KorNLI and KorSTS, two benchmark datasets for NLI and STS  in the Korean language.

\section{\ourbench~Evaluation}\label{appx:eval}
\begin{figure*}[]
\begin{centering}
  \includegraphics[width=0.85\textwidth]{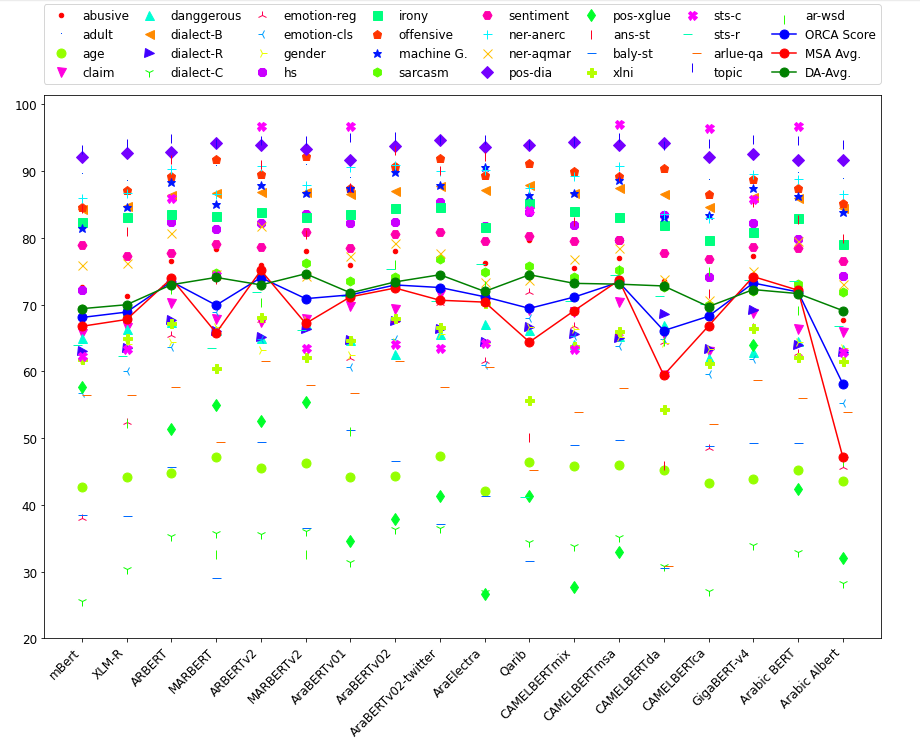}
  \caption{Models by ORCA score across all 29 tasks in~\ourbench~benchmark.} \label{appdx_fig:preformace_all_tasks}  \end{centering}
 \end{figure*}

\begin{figure*}[]
\begin{centering}
  \includegraphics[width=0.9\textwidth]{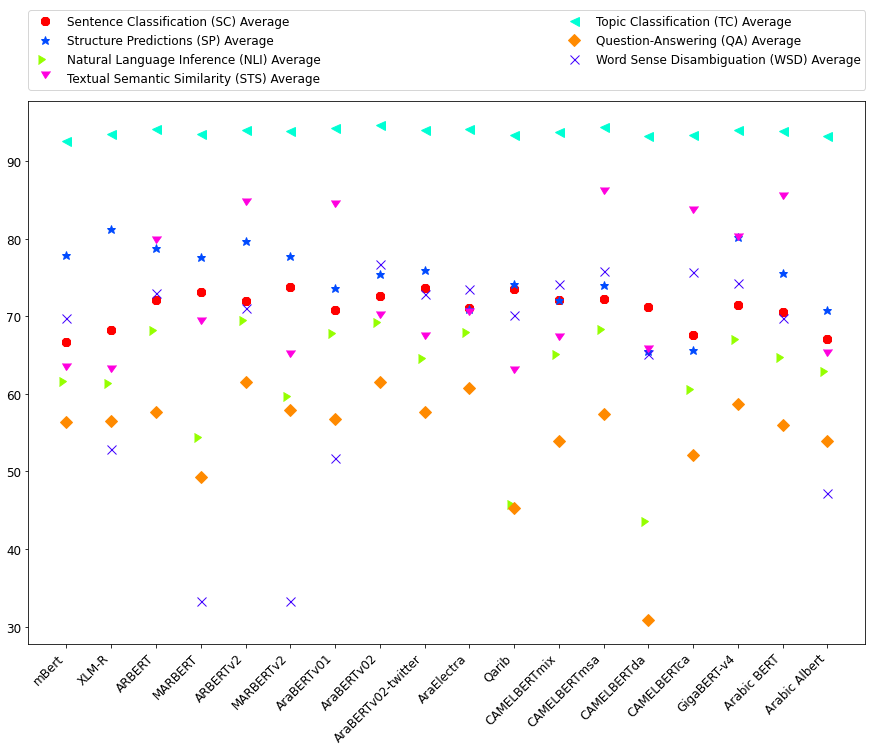}
  \caption{Models by ORCA score across all clusters in~\ourbench~benchmark.} \label{appdx_fig:preformace_all_clusters}  \end{centering}
 \end{figure*}

\begin{figure*}[]
\begin{centering}
  \includegraphics[width=0.85\textwidth]{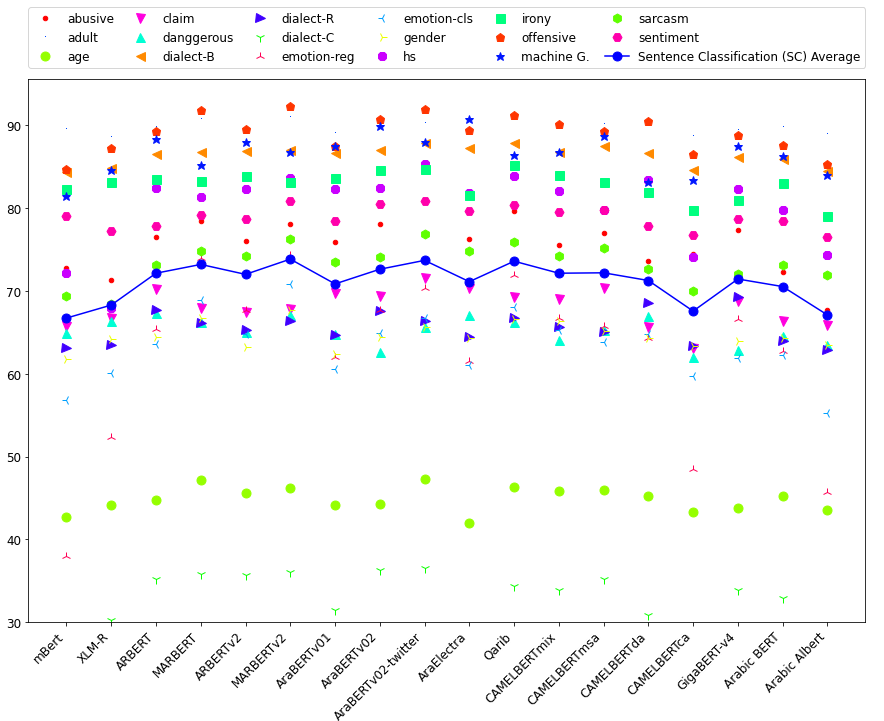}
  \caption{Models by F\textsubscript{1} score across all tasks in sentence classification cluster.} \label{appdx_fig:preformace_SC_clusters}  \end{centering}
 \end{figure*}

In this section, we provide additional information about the evaluation as follows: 
\begin{itemize}
    \item Performance of Arabic BERT-based models on \ourbench~Dev splits are shown in Table~\ref{tab:res_ar_LMs_dev}.
    \item Figure~\ref{appdx_fig:preformace_all_tasks} shows ORCA scores from the different PLMs across all $29$ tasks in~\ourbench~benchmark.
    \item Figure~\ref{appdx_fig:preformace_all_clusters} shows models' ORCA scores across all clusters in~\ourbench~benchmark.
    \item Figure~\ref{appdx_fig:preformace_SC_clusters} shows models' F\textsubscript{1} scores across all tasks in sentence classification cluster.
\end{itemize}



\begin{table*}[] 

\centering
 \renewcommand{\arraystretch}{1}
\resizebox{1\textwidth}{!}{%
\begin{tabular}{l@{\hspace{0.75\tabcolsep}}l@{\hspace{0.75\tabcolsep}}|l@{\hspace{0.75\tabcolsep}}l|@{\hspace{0.75\tabcolsep}}l@{\hspace{0.75\tabcolsep}}l@{\hspace{0.75\tabcolsep}}l@{\hspace{0.75\tabcolsep}}l@{\hspace{0.75\tabcolsep}}H@{\hspace{0.75\tabcolsep}}l@{\hspace{0.75\tabcolsep}}l@{\hspace{0.75\tabcolsep}}l@{\hspace{0.75\tabcolsep}}l@{\hspace{0.75\tabcolsep}}l@{\hspace{0.75\tabcolsep}}l@{\hspace{0.75\tabcolsep}}l@{\hspace{0.75\tabcolsep}}l@{\hspace{0.75\tabcolsep}}l@{\hspace{0.75\tabcolsep}}l@{\hspace{0.75\tabcolsep}}l@{\hspace{0.75\tabcolsep}}l@{\hspace{0.75\tabcolsep}}H}  
 \toprule  
                                      \textbf{\colorbox{blue!0}{Cluster}}  &
                                      \textbf{\colorbox{blue!0}{Task}}&
                                      \textbf{\colorbox{gray!10}{{B1}}}  &
                                      \textbf{\colorbox{gray!10}{{B2}}}&
                                      \textbf{\colorbox{blue!10}{{M1 }}} &
                                     \textbf{\colorbox{blue!10}{{M2}}}&
                                      \textbf{\colorbox{green!20}{{M3 }}} &
                                                     \textbf{\colorbox{green!20}{{M4   }}} &
                                                     \textbf{\colorbox{green!20}{{M5 }}}  & 
                                                     \textbf{\colorbox{red!10}{{M5 }}} &
                                                     \textbf{\colorbox{red!10}{{M6 }}} &
                                                     \textbf{\colorbox{red!10}{{M7 }}} & 
                                                     \textbf{\colorbox{red!10}{{M8 }}}& 
                                                     \textbf{\colorbox{orange!25}{{M9 }}}  & 
                                                     \textbf{\colorbox{yellow!25}{{M10 }}}  &
                                                     \textbf{\colorbox{yellow!25}{{M11 }}} &
                                                     \textbf{\colorbox{yellow!25}{{M12  }}}&
                                                     \textbf{\colorbox{yellow!25}{{M13  }}} & 
                                                     \textbf{\colorbox{red!35}{{M14 }}} &
                                                     \textbf{\colorbox{blue!25}{{M15 }}} &
                                                     \textbf{\colorbox{purple!35}{{M16 }}}  & 
                                                     \textbf{\colorbox{blue!10}{{M17 }}}  \\
                                                     \toprule 

\multicolumn{1}{l}{\multirow{11}{*}{\textbf{SC}}} 
&  abusive\textsuperscript{$\dagger$} &  $72.68$  &  $71.31$  &  $76.53$  &  $78.36$  &  $75.99$  &  $78.03$  &  $75.61$  &  $75.92$  &  $78.06$  &  $76.22$  &  $76.87$  &  $\bf79.66$  &  $75.49$  &  $76.98$  &  $73.57$  &  $74.43$  &  $77.34$  &  $72.28$  &  $67.68$ \\
& adult\textsuperscript{$\dagger$} & $89.52$ & $88.49$ & $89.7$ & $90.76$ & $89.67$ & $\bf90.97$ & $89.59$ & $88.97$ & $89.9$ & $89.65$ & $90.18$ & $90.89$ & $90.33$ & $90.09$ & $90.76$ & $88.68$ & $89.35$ & $89.74$ & $88.88$ \\
& age\textsuperscript{$\dagger$} & $42.68$ & $44.14$ & $44.76$ & $47.11$ & $45.57$ & $46.24$ & $44.93$ & $44.10$ & $44.33$ & $42.02$ & $\bf47.26$ & $46.35$ & $45.89$ & $45.97$ & $45.29$ & $43.29$ & $43.83$ & $45.23$ & $43.61$ \\
&  claim$^\star$ &  $65.72$  &  $66.66$  &  $70.25$  &  $67.91$  &  $67.38$  &  $67.83$  &  $68.01$  &  $69.74$  &  $69.34$  &  $70.35$  &  $\bf71.53$  &  $69.2$  &  $68.96$  &  $70.32$  &  $65.66$  &  $63.06$  &  $68.81$  &  $66.29$  &  $65.88$ \\
 & dangerous\textsuperscript{$\dagger$}  &  $64.94$  &  $66.31$  &  $\bf67.32$  &  $66.2$  &  $64.96$  &  $67.11$  &  $63.48$  &  $64.72$  &  $62.6$  &  $67.13$  &  $65.66$  &  $66.25$  &  $64.03$  &  $65.31$  &  $66.92$  &  $61.97$  &  $62.83$  &  $64.56$  &  $63.41$ \\
& dialect-b\textsuperscript{$\dagger$}& $84.29$ & $84.78$ & $86.48$ & $86.78$ & $86.92$ & $86.91$ & $94.32$ & $86.64$ & $87.01$ & $87.76$ & $87.21$ & $\bf87.85$ & $86.79$ & $87.40$ & $86.64$ & $84.58$ & $86.57$ & $86.13$ & $85.94$ \\										
&  dialect-r\textsuperscript{$\dagger$} &  $63.12$  &  $63.51$  &  $67.71$  &  $66.08$  &  $65.21$  &  $66.32$  &  $65.29$  &  $64.63$  &  $67.5$  &  $64.46$  &  $66.34$  &  $66.71$  &  $65.59$  &  $65.05$  &  $68.55$  &  $63.36$  &  $\bf69.22$  &  $63.98$  &  $62.87$ \\
&  dialect-c\textsuperscript{$\dagger$}  &  $25.52$  &  $30.34$  &  $35.26$  &  $35.83$  &  $35.69$  &  $36.06$  &  $36.42$  &  $31.49$  &  $36.33$  &  $27.00$  &  $\bf36.50$  &  $34.36$  &  $33.90$  &  $35.18$  &  $30.83$  &  $27.05$  &  $33.96$  &  $32.99$  &  $28.25$ \\
&  emotion\textsuperscript{$\dagger$} & $56.79$ & $60.05$ & $63.6$ & $68.85$ & $64.81$ & $\bf70.82$ & $63.21$ & $60.6$ & $64.89$ & $60.98$ & $66.70$ & $68.03$ & $65.25$ & $63.85$ & $64.8$ & $59.66$ & $61.92$ & $62.2$ & $55.22$ \\
& emotion-reg $^\star$ & $37.96$ & $52.37$ & $65.37$ & $73.96$ & $67.73$ & $\bf74.27$ & $67.83$ & $62.02$ & $67.64$ & $61.51$ & $70.31$ & $71.91$ & $66.73$ & $65.75$ & $64.34$ & $48.46$ & $66.57$ & $62.77$ & $45.72$ \\
& gender\textsuperscript{$\dagger$} & $61.78$ & $64.16$ & $64.38$ & $66.65$ & $63.18$ & $\bf67.64$ & $63.51$ & $62.41$ & $64.37$ & $64.24$ & $65.65$ & $66.64$ & $66.38$ & $65.19$ & $64.25$ & $63.37$ & $63.97$ & $64.35$ & $63.50$ \\

 &  hate\textsuperscript{$\dagger$}  &  $72.19$  &  $67.88$  &  $82.41$  &  $81.33$  &  $82.26$  &  $83.54$  &  $80.66$  &  $82.21$  &  $82.39$  &  $81.79$  &  $\bf85.30$  &  $83.88$  &  $81.99$  &  $79.68$  &  $83.38$  &  $74.1$  &  $82.25$  &  $79.77$  &  $74.26$ \\
  &  irony\textsuperscript{$\dagger$}  &  $82.31$  &  $83.13$  &  $83.53$  &  $83.27$  &  $83.83$  &  $83.09$  &  $84.12$  &  $83.63$  &  $84.51$  &  $81.56$  &  $84.62$  &  $\bf85.16$  &  $84.01$  &  $83.07$  &  $81.91$  &  $79.68$  &  $80.91$  &  $83.03$  &  $79.05$ \\
   &  offensive\textsuperscript{$\dagger$}  &  $84.62$  &  $87.18$  &  $89.28$  &  $91.84$  &  $89.55$  &  $\bf92.23$  &  $90.14$  &  $87.5$  &  $90.73$  &  $89.4$  &  $91.89$  &  $91.17$  &  $90.05$  &  $89.32$  &  $90.44$  &  $86.52$  &  $88.76$  &  $87.52$  &  $85.26$ \\

  &  machine G.$^\star$  &  $81.4$  &  $84.61$  &  $88.35$  &  $85.14$  &  $87.94$  &  $86.69$  &  $87.69$  &  $87.45$  &  $89.82$  &  $\bf90.66$  &  $87.96$  &  $86.35$  &  $86.73$  &  $88.62$  &  $83.17$  &  $83.35$  &  $87.43$  &  $86.28$  &  $83.91$ \\
  
  &  sarcasm\textsuperscript{$\dagger$}  &  $69.32$  &  $68.42$  &  $73.11$  &  $74.74$  &  $74.16$  &  $76.19$  &  $75.35$  &  $73.46$  &  $74.06$  &  $74.81$  &  $\bf76.83$  &  $75.82$  &  $74.17$  &  $75.18$  &  $72.57$  &  $69.94$  &  $72.02$  &  $73.11$  &  $71.92$ \\

& sentiment\textsuperscript{$\dagger$} & $78.99$ & $77.21$ & $77.78$ & $79.08$ & $78.60$ & $80.83$ & $78.85$ & $78.45$ & $80.50$ & $79.56$ & $\bf80.86$ & $80.33$ & $79.51$ & $79.75$ & $77.8$ & $76.76$ & $78.68$ & $78.46$ & $76.46$ \\\hdashline


\multicolumn{1}{l}{\multirow{2}{*}{\textbf{NER}}} 
 &  anerc.  &  $85.92$  &  $86.76$  &  $90.27$  &  $86.59$  &  $90.83$  &  $87.86$  &  $90.87$  &  $90.68$  &  $\bf90.85$  &  $90.17$  &  $90.03$  &  $87.5$  &  $89.27$  &  $90.71$  &  $83.61$  &  $82.94$  &  $89.52$  &  $88.77$  &  $86.54$ \\
  &  aqmar  &  $75.95$  &  $76.16$  &  $80.72$  &  $74.57$  &  $\bf81.70$  &  $74.22$  &  $80.19$  &  $77.34$  &  $79.2$  &  $73.43$  &  $77.66$  &  $73.72$  &  $76.84$  &  $78.54$  &  $73.77$  &  $70.71$  &  $74.97$  &  $79.5$  &  $73.15$ \\
  & pos-dia\textsuperscript{$\dagger$} & $92.04$ & $92.78$  &  $92.92$  &  $94.14$  &$93.92$  & $93.38$  &   xx  &  $91.65$  &  $93.79$  &  $94.70$   & $93.554$ & $\bf94.70$ & $93.95$ & $94.37$ & $93.95$ & $92.05$ & $92.05$ & $93.24$ & $92.57$ &  $xx$ \\ 
   &  pos-xglue\textsuperscript{$\star$}    &   $57.68$ & $69.37$ & $51.39$ & $55.02$ & $52.55$ & $55.45$ & $34.65$&  $34.65$ & $37.84$ & $41.28$ & $26.61$ & $41.36$ & $27.70$ & $32.89$ & $10.37$ & $17.04$ & $62.58$ & $\bf63.89$  & $42.40$ & $32.10$  \\ \hline
  
 \multicolumn{1}{l}{\multirow{3}{*}{\textbf{NLI}}} 

 &  ans-st\textsuperscript{$\star$}\textsuperscript{$\ddagger$}  &  $84.49$  &  $81.00$  &  $91.77$  &  $73.82$  &  $91.02$  &  $80.57$  &  $88.87$  &  $87.59$  &  $\bf93.23$  &  $92.33$  &  $90.17$  &  $50.2$  &  $82.49$  &  $89.21$  &  $46.01$  &  $71.81$  &  $85.31$  &  $82.86$  &  $80.02$ \\
 &  baly-st\textsuperscript{$\star$}  &  $34.48$  &  $38.27$  &  $45.63$  &  $29.07$  &  $49.34$  &  $36.52$  &  $49.82$  &  $\bf51.19$  &  $46.63$  &  $41.32$  &  $37.12$  &  $31.58$  &  $48.94$  &  $49.67$  &  $30.58$  &  $48.85$  &  $49.21$  &  $49.22$  &  $47.19$ \\ 
  &  xlni\textsuperscript{$\star$}  &  $61.88$  &  $65.06$  &  $67.22$  &  $60.50$  &  $68.17$  &  $62.22$  &  $67.95$  &  $64.69$  &  $67.93$  &  $\bf70.20$  &  $66.67$  &  $55.67$  &  $63.82$  &  $66.02$  &  $54.29$  &  $61.18$  &  $66.53$  &  $62.15$  &  $61.62$ \\\hline 
\multirow{2}{*}{\textbf{STS}}  

  &  sts-r$^{\dagger}$  &  $63.91$  &  $62.24$  &  $73$  &  $63.48$  &  $71.90$  &  $66.12$  &  $73.19$  &  $71.27$  &  $75.4$  &  $\bf76.01$  &  $70.50$  &  $41.15$  &  $70.61$  &  $74.42$  &  $71.23$  &  $70.13$  &  $73.68$  &  $73.56$  &  $66.75$ \\
  &  sts-c\textsuperscript{$\star$}  &  $62.34$  &  $63.35$  &  $85.95$  &  $74.43$  &  $96.73$  &  $63.47$  &  $85.98$  &  $96.81$  &  $64.11$  &  $64.24$  &  $63.52$  &  $84.11$  &  $63.28$  &  $\bf97.10$  &  $59.57$  &  $96.41$  &  $85.87$  &  $96.69$  &  $62.91$ \\\hline 
  \textbf{TC} &  
  topic  &  $92.55$  &  $93.53$  &  $94.17$  &  $93.53$  &  $93.96$  &  $93.9$  &  $94.32$  &  $94.31$  &  $\bf94.58$  &  $94.11$  &  $94.02$  &  $93.32$  &  $93.72$  &  $94.38$  &  $93.18$  &  $93.41$  &  $94.05$  &  $93.86$  &  $93.27$ \\ \hline

 \multirow{1}{*}{\textbf{QA}}  
   &  arlue-qa &  $56.39$  &  $56.51$  &  $57.65$  &  $49.35$  &  $61.5$  &  $57.9$  &  $61.46$  &  $56.79$  &  $\bf61.56$  &  $60.70$  &  $57.65$  &  $45.27$  &  $53.98$  &  $57.46$  &  $30.91$  &  $52.11$  &  $58.71$  &  $55.94$  &  $53.89$ \\ 
    
    & pos-dia\textsuperscript{$\dagger$} & $92.04$ & $92.78$  &  $92.92$  &  $94.14$  &$93.92$  & $93.38$  &   xx  &  $91.65$  &  $93.79$  &  $94.70$   & $93.554$ & $\bf94.70$ & $93.95$ & $94.37$ & $93.95$ & $92.05$ & $92.05$ & $93.24$ & $92.57$ &  $xx$ \\

     \hline

 \colorbox{red!10}{\textbf{Avg. Dia }} &  & $67.76$ & $68.35$& $71.56$ & $72.63$ & $71.45$ & \colorbox{cyan!10}{$\bf73.28$} & $71.82$ & $70.33$ & $71.94$ & $70.47$ & $72.99$& \underline{$73.07$} & $71.67$ & $71.57$& $71.26$ & $68.09$ & $70.82$& $70.23$ & $67.59$  \\   \hline 
     \toprule  
 \colorbox{green!10}{\textbf{Avg. MSA }} &   &$66.91$ & $68.87$ & $75.86$ & $69.36$ & \colorbox{green!10}{$\bf77.35$} & $70.90$ & $76.34$& $75.82$ & $75.02$ & $73.75$ & $73.09$ & $65.83$ & $72.11$ & \underline{$76.85$} & $63.02$ & $70.20$ & $75.05$ & $74.82$& $68.40$ \\ \hline

\multicolumn{2}{l}{\colorbox{cyan!25}{\textbf{~~~~\ourbench\textsubscript{$score$}}}}&  $67.34$ & $68.61$ & $73.71$ &$70.99$ & \colorbox{cyan!25}{$\bf74.40$} &$72.12$ & $74.08$ & $73.08$ & $73.48$ & $72.11$ & $73.04$ & $69.45$ & $71.89$ & \underline{$74.21$} & $67.14$ & $69.15$ & $72.94$ & $72.53$ & $67.99$ \\

\toprule  

\end{tabular}} \caption{\small{Performance of  Arabic Bert-based  models on \ourbench~Dev  splits (F$_1$).    $^{\ddagger}$~Metric for STSP taks is spearman correlation.  \colorbox{gray!15}{\textbf{B1, B2}:} Two baselines mBERT~\cite{devlin2019bert} and XLM-R~\cite{liu2019roberta}. \colorbox{blue!10}{\textbf{M1, M2}:}  ARBERT, MARBERT~\cite{abdul2020arbert}. \colorbox{green!20}{\textbf{M3, M4}:} ARBERT\textsubscript{V2} and MARBERT\textsubscript{V2}.  \colorbox{red!10}{\textbf{M5, M6, M7, and M8}:}  AraBERT\textsubscript{v1[v2, tw]}, and AraElectra~\cite{antoun2020arabert, antoun2020araelectra}.  \colorbox{orange!25}{\textbf{M9}:} Qraib~\cite{chowdhury2020} \colorbox{yellow!25}{\textbf{M10, M11, M12, and M13}:} CamelBERT\textsubscript{mix[msa, da, ca]}~\cite{CAMeLBERT2021}.  \colorbox{red!35}{\textbf{M14}:} GigaBERT\textsubscript{v4}~\cite{chowdhury2020}.  \colorbox{blue!25}{\textbf{M15}:} Arabic BERT~\cite{chowdhury2020}. \colorbox{purple!25}{\textbf{M16}}: Arabic Albert~\cite{lan2020gigabert}. \textbf{Avg. Dia, and Avg. MSA:}~The average of dialect and MSA tasks. \colorbox{cyan!25}{\textbf{\ourbench\textsubscript{$score$}}}: Average overall Dia and MSA tasks. \textsuperscript{$\star$}MSA tasks. \textsuperscript{$\ddagger$}DIA tasks.   A task is  considered  as an  MSA if it has more than $98$\% samples  predicted as MSA using an MSA Vs DIA classifier (see Table~\ref{tab:SuperARLUE_msa_dia}).} }
\label{tab:res_ar_LMs_dev}

\end{table*}

\section{Public leaderboard.}\label{apdx:leaderboard}

\begin{figure*}[]
\begin{centering}
  \includegraphics[width=0.95\textwidth]{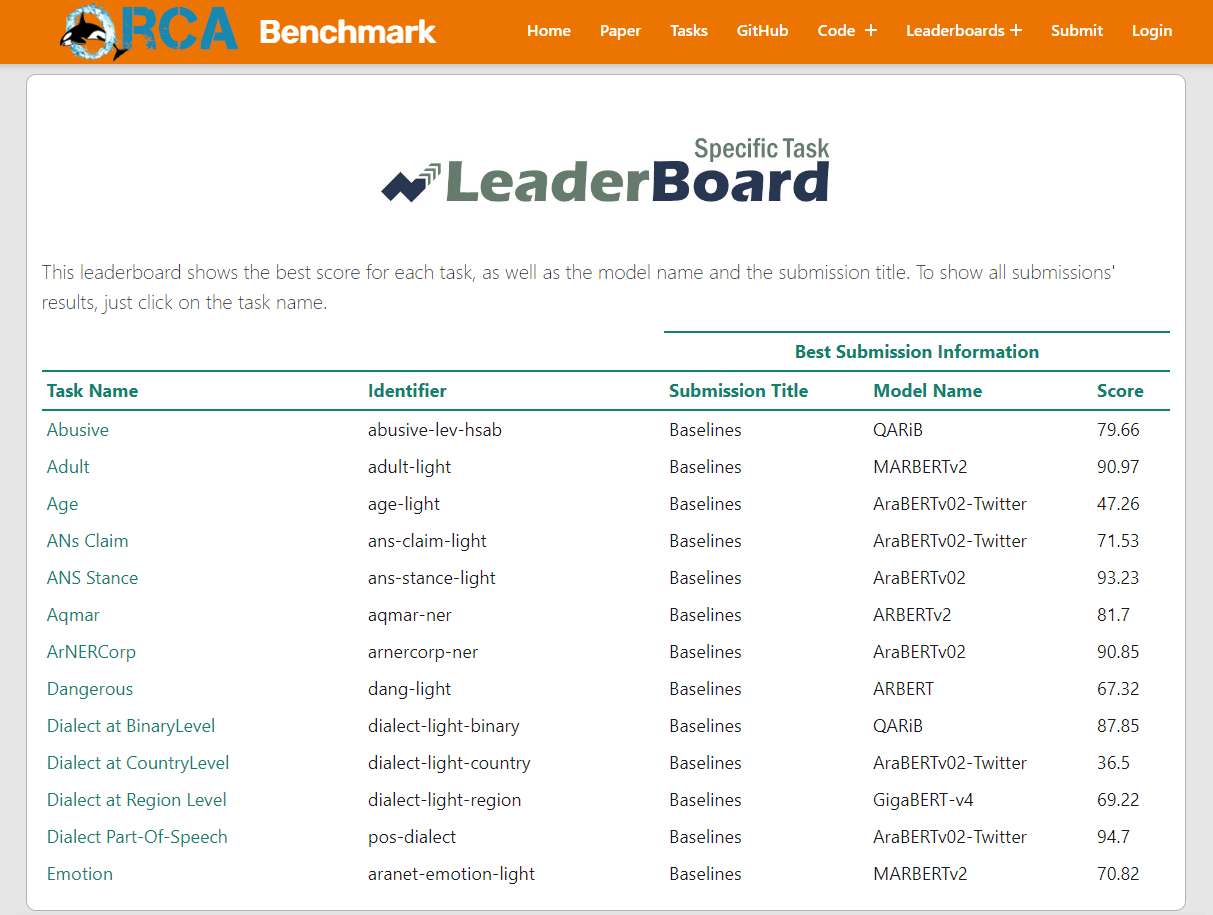}
  \caption{OCRA leaderboard for example tasks sorted alphabetically.} \label{fig:screen_onetask}  \end{centering}
 \end{figure*}

In this section, we provide additional screenshots for~\ourbench leaderboard, as follows: 
\begin{itemize}
    \item Figure~\ref{fig:screen_onetask} shows an example of tasks sorted alphabetically.
    \item Figure~\ref{fig:screen_tasks_leaderboard} shows detailed scores by all models for a given task..
    \item Figure~\ref{fig:screen_tasks_leaderboard} shows detailed information about each task cluster and associated tasks, with each task assigned an identifier, language variety, evaluation metric, a link to the dataset website/GitHub/paper and bibliographic information.
\end{itemize}
 
  \begin{figure*}[]
\begin{centering}
  \includegraphics[width=0.9\textwidth]{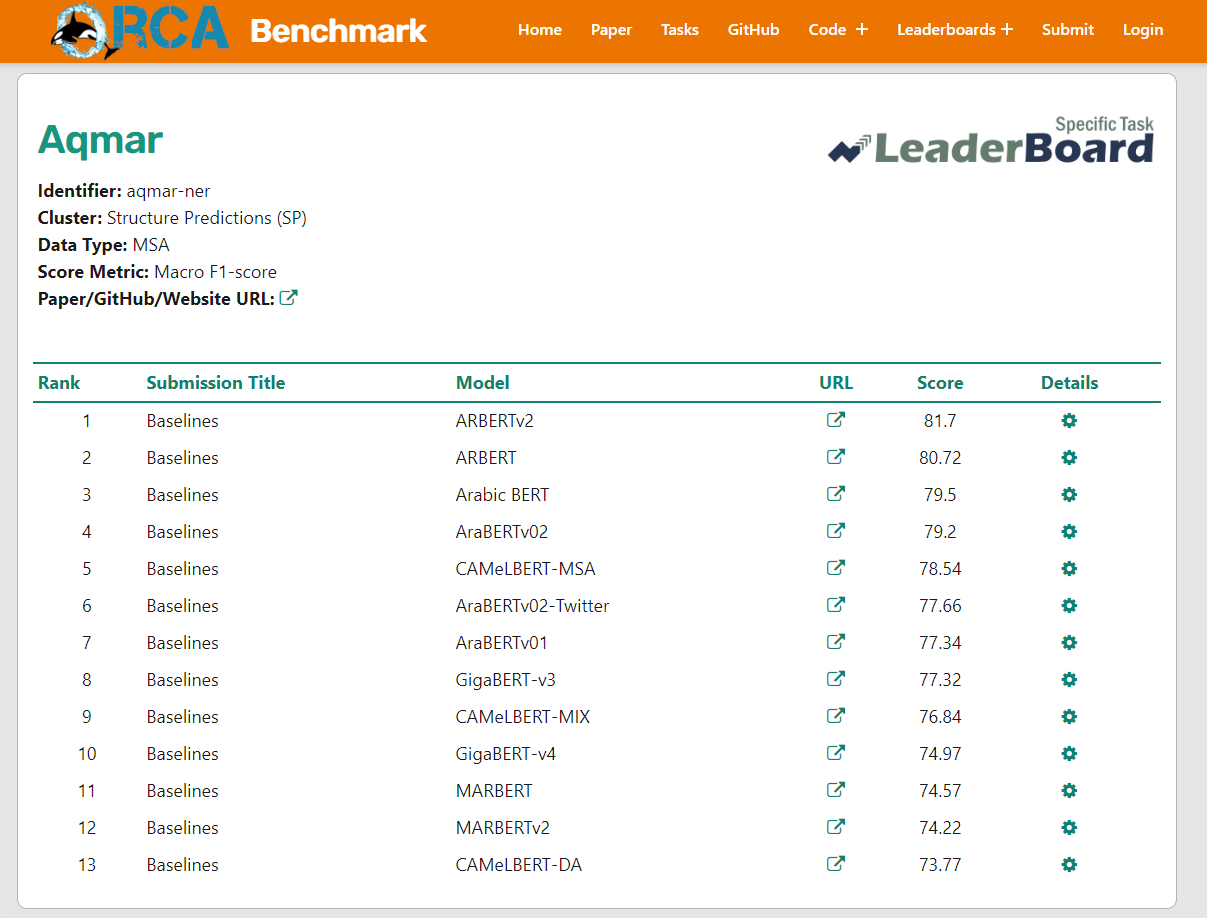}
  \caption{Modularity of OCRA leaderboard allows showing detailed scores by all models for a given task.} \label{fig:screen_tasks_leaderboard}  \end{centering}
 \end{figure*}
 
   \begin{figure*}[]
\begin{centering}
  \includegraphics[width=0.9\textwidth]{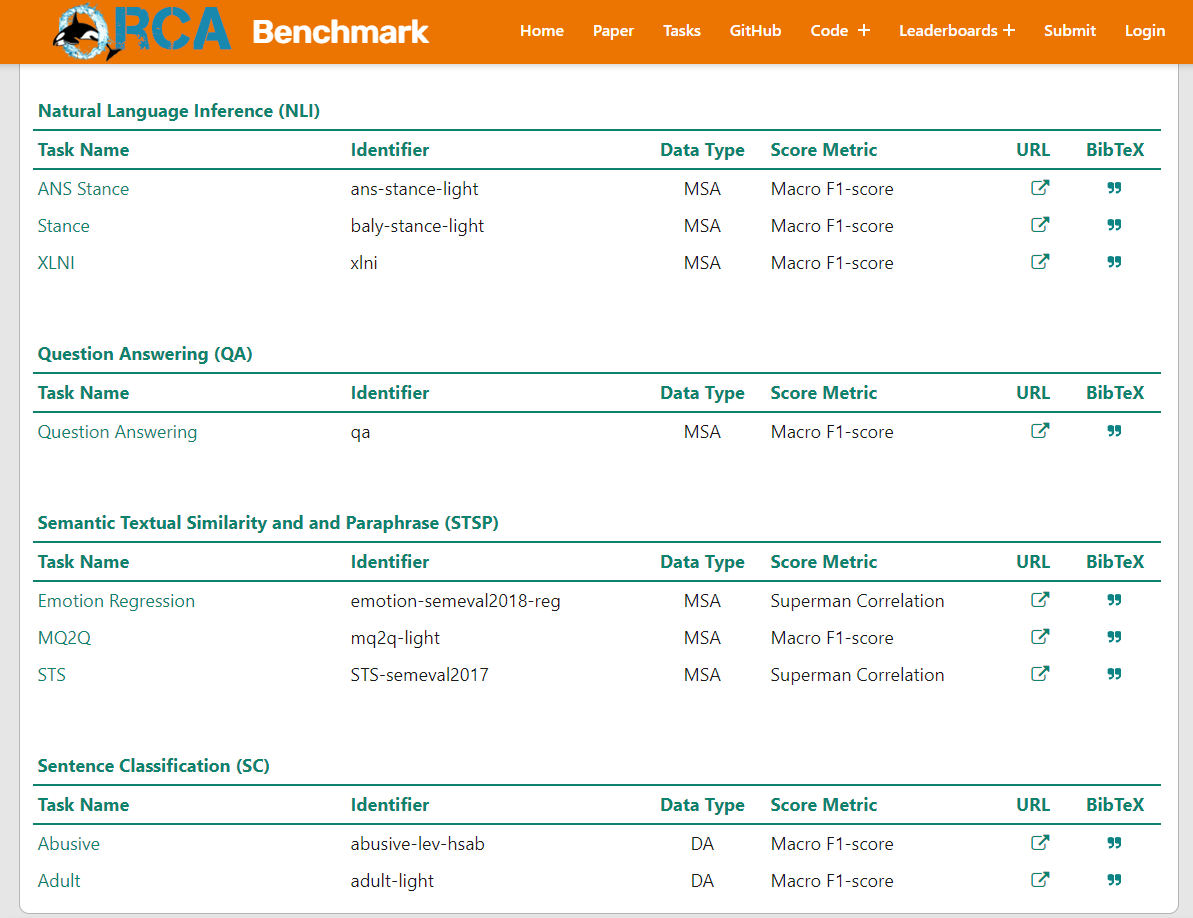}
  \caption{OCRA leaderboard also provides detailed information about each task cluster and associated tasks, with each task assigned an identifier, language variety, evaluation metric, a link to the dataset website/GitHub/paper and  bibliographic information.} \label{fig:screen_tasks_leaderboard}  \end{centering}
 \end{figure*}

 \section{Analysis of Model Computational Cost}\label{appx_sec_model_analysis}

\begin{figure*}[t]
 \centering
 \includegraphics[scale=0.48]{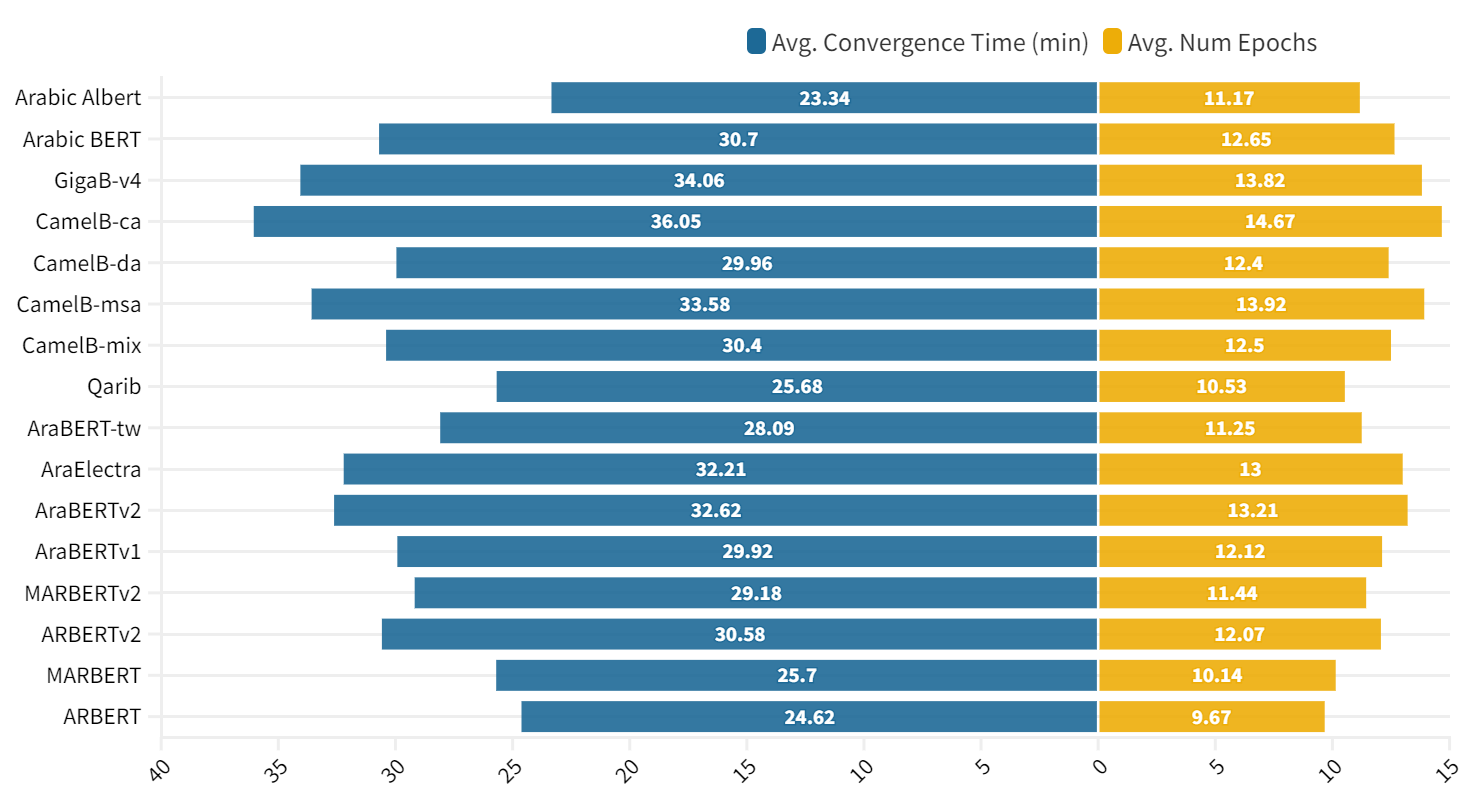}
  \caption{The average number of epochs (in orange),  and time needed to converge (mins, in blue) for all the studied pretrained language  models across all \ourbench~tasks.} \label{fig:avg_conv} 
 \end{figure*}

\begin{figure*}[t]
  \centering
 \includegraphics[width=\textwidth]{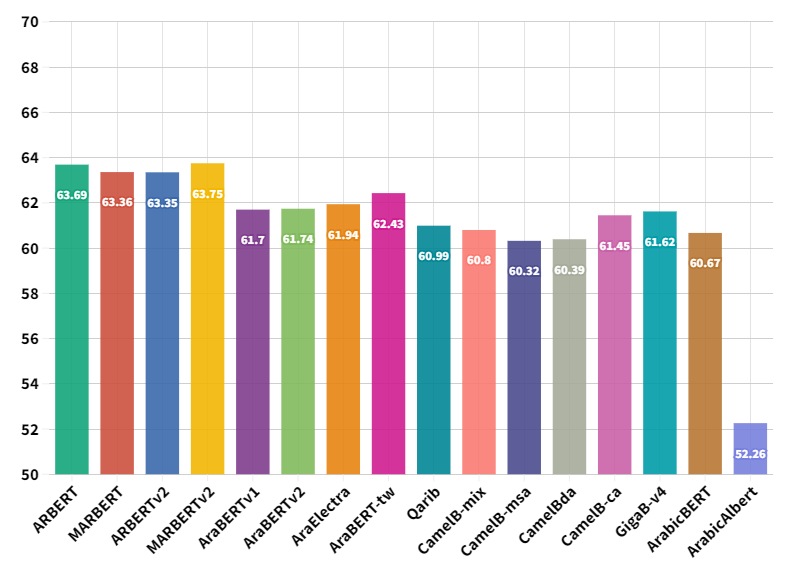}
  \caption{The time needed in minutes to fine-tune ($25$ epochs). We compute the average time of three runs  across all 
  \ourbench~tasks.} \label{fig:25_epoch} 
 \end{figure*}

In this section, we provide additional information about the models' computational cost, as follows: 
\begin{itemize}
    \item Figure~\ref{fig:avg_conv} shows the average number of epochs (in orange),  and time needed to converge (mins, in blue) for all the studied pretrained language  models across all~\ourbench~tasks.
    \item Figure~\ref{fig:25_epoch} shows the time needed in minutes to fine-tune ($25$ epochs). We compute the average time of three runs  across all 
  \ourbench~tasks.
\end{itemize}
  
 \section{\ourbench~Data}\label{appx_sec_model_data}

\begin{figure*}[t]
\begin{centering}
  \frame{\includegraphics[width=0.9\textwidth]{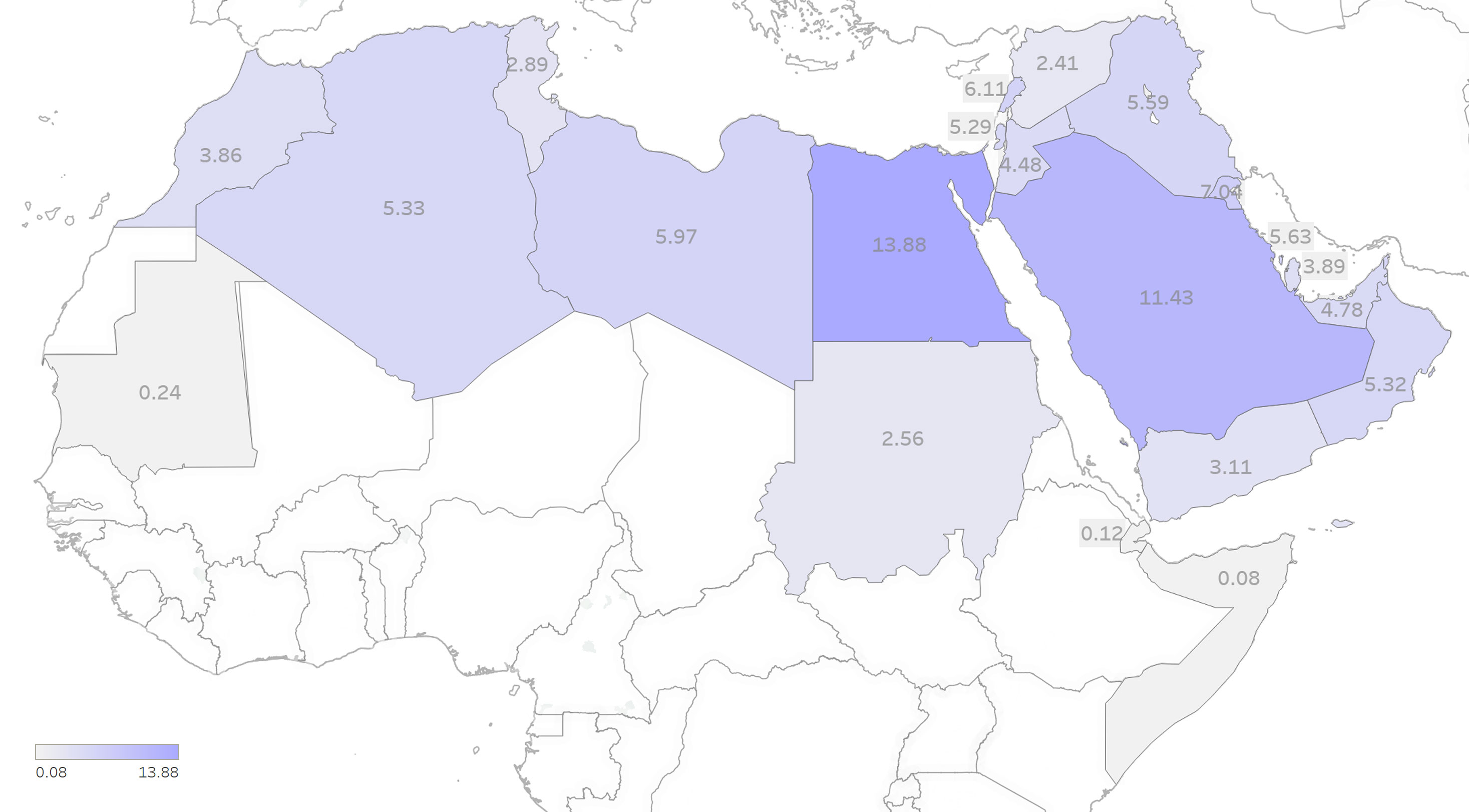}}
  \caption{Predicted country-level distribution,  in percentage, in the dialectal portion of \ourbench.} \label{fig:map_countery}  \end{centering}
 \end{figure*}

In this section, we provide additional information about~\ourbench Data, as follows: 
\begin{itemize}
    \item Table~\ref{tab:samples} shows a randomly picked examples from the dialectal portion of~\ourbench~Train datasets.
    \item Figure~\ref{fig:map_countery} shows the predicted country-level distribution,  in percentage, in the dialectal portion of \ourbench.
\end{itemize}
  \begin{table*}[!ht]
\centering
 \renewcommand{\arraystretch}{1.2}
\resizebox{\textwidth}{!}{%
\begin{tabular}{lrll}
\toprule
\multicolumn{1}{c}{\textbf{Country}} & \multicolumn{1}{c}{\textbf{Example}}                                              & \multicolumn{1}{c}{\textbf{Dataset~~~~~~}} & \textbf{Label}\\ \midrule
\multirow{3}{*}{Egypt}                 &   <ايطاليا و انجلترا خيبوا توقعاتى بس> 
                                         & Emotion                                 &  Happy\\ 
  & <لن يفهمك، فأنت تتحدث عن أمر قطعت فيه آلاف الأميال تفكيرا ولم يمش فيه خطوة >                                        & Adult                             &  Not Adult\\ 
    & <الخلفة الوسخة بتجيب لاهلها التهزيق> & Sarcasm                                &  Sarcasm\\ \hline

\multirow{3}{*}{Jordan}                & 
< ايوون لازم انزله هاد عشان بس افوز اجمع سكور بس بعدك ضعيفه انا 20 الف >                             & Gender                    &  Male\\ 
& <ومن وراكي يا نشميه يا ام رح تتطبق علينا >                            & Offensive                               &  Not Offensive\\ 
      & < ما احد يربط هالجحش>                           & Abusive                                &  Abusive \\  \hline
\multirow{3}{*}{KSA}                   & <اذا تسوين شى	>                                              & Dangerous                              &  Dangerous\\ 
             & <وش رايكم تحذفون الاغاني وتحطون ايديكم على قلوبكم !>                                        & Emotion                            &  Happy\\ 
      & <اكره الي تصير مشرفة دعم متسابق وتوكل نفسها محامية وتجيك ٢٤ ساعه تراقب التايم     >    & Age                               &  Under 25\\  \hline
\multirow{3}{*}{Kuwait}                & <للاسف الشبكه تعيسه جدا لا بديراب ولا بالعماريه والعيينه ولا بشقرا      	>                      & Sentiment            &  Negative\\ 
& <نفسي . كل مره احط اليوز و الرقم ولمه ادشه بعد ٥ دقايق القاه طالع >        & Gender                     &  Male\\ 
  & <هه عااد ماله امان هوو كلش اي علي اسم عيالتكم هه	>                                     & Emotion                                 &  Fear\\  
  
  \bottomrule
\end{tabular}%
}
\caption{Randomly picked examples from the dialectal portion of \ourbench~Train datasets.}\label{tab:samples}
\end{table*}

\end{document}